\documentclass[onefignum,onetabnum]{siamart171218}
\usepackage{graphicx}
\usepackage{epstopdf}
\usepackage{bm}
\usepackage{amssymb}
\usepackage{psfrag}
\usepackage{color}
\usepackage{amsbsy}      
\usepackage{amsmath}
\usepackage{mathtools}
\usepackage{dsfont}
\usepackage{comment}
\usepackage{hyperref}
\usepackage{hhline}
\usepackage{tcolorbox}
\usepackage{makecell}
\usepackage{booktabs}
\usepackage{float,lscape}
\usepackage{dsfont}
\usepackage{url}
\usepackage{braket}
\usepackage{etoolbox}
\usepackage{soul, color}
\usepackage{amsopn}
\def\ie{\textit{i.e.}}
\def\eg{\textit{e.g.}}
\DeclareMathOperator*{\argmin}{arg\,min}

\ifpdf
\hypersetup{
  pdftitle={Near-optimal control of dynamical systems with neural ordinary differential equations},
  pdfauthor={Lucas B\"ottcher and Thomas Asikis}
}
\fi

\usepackage{lipsum}
\usepackage{amsfonts}
\usepackage{graphicx}
\usepackage{epstopdf}
\usepackage{algorithmic}
\ifpdf
  \DeclareGraphicsExtensions{.eps,.pdf,.png,.jpg}
\else
  \DeclareGraphicsExtensions{.eps}
\fi

\newsiamremark{remark}{Remark}
\newsiamremark{hypothesis}{Hypothesis}
\crefname{hypothesis}{Hypothesis}{Hypotheses}
\newsiamthm{claim}{Claim}

\headers{Near-optimal control with neural ODE{$\mathrm{s}$}}{L. B\"ottcher and T. Asikis}

\title{Near-optimal control of dynamical systems with neural
ordinary differential equations\thanks{Submitted to the editors on June 10, 2022.
\funding{Lucas B\"ottcher thanks Mingtao Xia for helpful discussions. Thomas Asikis acknowledges financial support from NCCR Automation. Both authors would like to especially thank Nino Antulov-Fantulin for the helpful inputs, discussion, and guidance.}}}

\author{Lucas B\"ottcher\thanks{Centre for Human and Machine Intelligence, Frankfurt School of Finance and Management, 60322 Frankfurt am Main, Germany
  (\email{l.boettcher@fs.de}).}
  \and Thomas Asikis\thanks{Game Theory, University of Zurich, 8006 Zurich, Switzerland
  (\email{thomas.asikis@uzh.ch}).}
    }
  
\usepackage{amsopn}

\makeatletter
\newcommand*{\addFileDependency}[1]{
  \typeout{(#1)}
  \@addtofilelist{#1}
  \IfFileExists{#1}{}{\typeout{No file #1.}}
}
\makeatother

\begin{document}

\maketitle

\begin{abstract}
Optimal control problems naturally arise in many scientific applications where one wishes to steer a dynamical system from a certain initial state $\mathbf{x}_0$ to a desired target state $\mathbf{x}^*$ in finite time $T$. Recent advances in deep learning and neural network\textendash based optimization have contributed to the development of methods that can help solve control problems involving high-dimensional dynamical systems. In particular, the framework of neural ordinary differential equations (neural ODEs) provides an efficient means to iteratively approximate continuous time control functions associated with analytically intractable and computationally demanding control tasks. Although neural ODE controllers have shown great potential in solving complex control problems, the understanding of the effects of hyperparameters such as network structure and optimizers on learning performance is still very limited. Our work aims at addressing some of these knowledge gaps to conduct efficient hyperparameter optimization. To this end, we first analyze how truncated and non-truncated backpropagation through time affect runtime performance and the ability of neural networks to learn optimal control functions. Using analytical and numerical methods, we then study the role of parameter initializations, optimizers, and neural-network architecture. Finally, we connect our results to the ability of neural ODE controllers to implicitly regularize control energy.
\end{abstract}

\begin{keywords}
dynamical systems, neural ODEs, optimal control, optimization, implicit regularization
\end{keywords}

\begin{AMS}
93B47, 93-08, 37N35, 65P99, 68T07
\end{AMS}
\date{\today}
\section{Introduction}
The optimal control (OC) of complex dynamical systems is relevant in many scientific disciplines~\cite{bertsekas2012dynamic}, including biology~\cite{jarrah2004optimal,lenhart2007optimal,laubenbacher2013agent,ledzewicz2016optimal,konstorum2017addressing,an2017optimization}, epidemiology~\cite{choi2021optimal,sharomi2017optimal}, quantum engineering~\cite{mabuchi2009continuous,dong2010quantum}, and power systems~\cite{minciardi2011optimal,bienstock2011optimal}.

Mathematically, optimal control is concerned with finding a control signal ${\bf u}(t)\in \mathbb{R}^m$ ($0\leq t\leq T$) that steers a dynamical system from its initial state ${\bf x}_0\in\mathbb{R}^n$ to a desired target state ${\bf x}^*\in\mathbb{R}^n$ in finite time $T$, minimizing the control energy 
\begin{equation}
E_T[{\bf u}]=\frac{1}{2}\int_0^T \|{\bf u}(t)\|_2^2\,\mathrm{d}t\,.
\label{eq:control_energy}
\end{equation}
Depending on the application, it may be useful to consider integrated cost functionals that are different from Eq.~\eqref{eq:control_energy}. A corresponding example is provided in Appendix~\ref{app:moving_particle}. 

The outlined optimal control problem aims at finding
\begin{equation}
{\bf u}^*(t)=\argmin_{{\bf u}(t)}E_T[{\bf u}]\,,
\label{eq:u_min}
\end{equation}
subject to the constraint
\begin{equation}
\dot{{\bf x}}={\bf f}({\bf x}(t),{\bf u}(t),t)\,,\quad {\bf x}(0)={\bf x}_0\,,\quad {\bf x}(T)={\bf x}^*\,,
\label{eq:dyn_sys}
\end{equation}
where $\mathbf{f}(\cdot)$ denotes a vector field that depends on the system state ${\bf x}(t)$, control input ${\bf u}(t)$, and time $t$. Linear systems admit a closed-form solution of ${\bf u}^*(t)$~\cite{yan2012controlling}. For general, non-linear dynamical systems, one typically aims at minimizing
\begin{equation}
J=L({\bf x}(T),{\bf x}^*)+\mu E_T[{\bf u}]\,,
\label{eq:oc_loss}
\end{equation}
where $L({\bf x}(T),{\bf x}^*)$ denotes the distance between the reached state and the desired target state ${\bf x}^*$, \eg, the mean-squared error (MSE) $L({\bf x}(T),{\bf x}^*)\propto\|{\bf x}(T)-{\bf x}^*\|_2^2$. The Lagrange multiplier $\mu$ models the impact of control energy on the total loss. Clearly, in the limit $\mu\rightarrow \infty$, we have $\mathbf{u}^*=0$.
\begin{figure}
    \centering
    \includegraphics[width=0.9\textwidth]{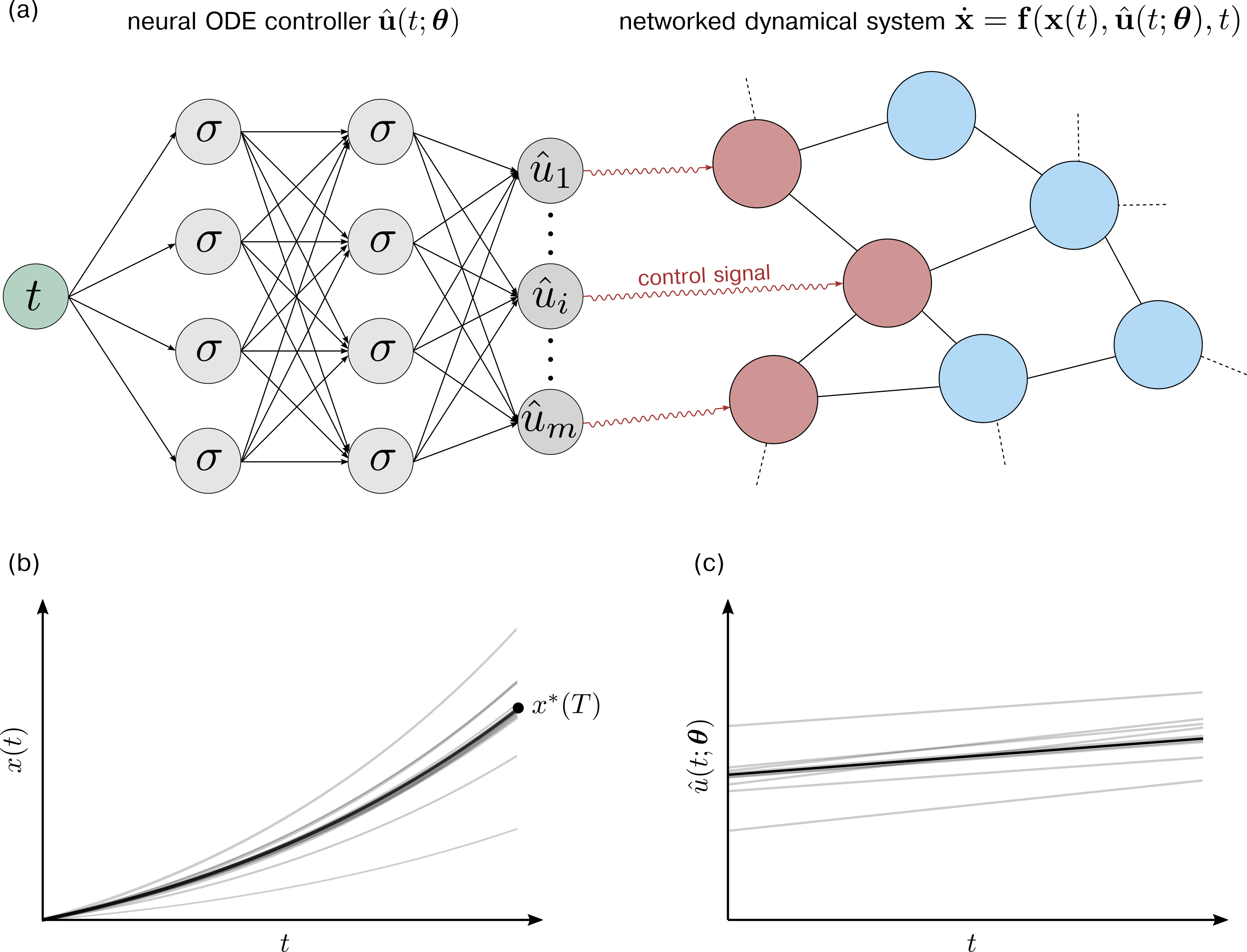}
    \caption{Controlling networked dynamical systems with neural ODEs. (a) A neural ODE controller takes the time $t$ as an input variable and produces a control signal $\hat{\mathbf{u}}(t;\boldsymbol{\theta})\in\mathbb{R}^m$. A networked dynamical system is then controlled by connecting control inputs $u_1(t;\boldsymbol{\theta}),\dots,u_m(t;\boldsymbol{\theta})$ to all or a subset of nodes. Activation functions in the neural ODE controller are denoted by $\sigma$. (b,c) Evolution of $x(t)$, the state of a one-dimensional dynamical system, and of $\hat{u}(t;\boldsymbol{\theta})$, the corresponding control function. Solid grey lines represent $x(t)$ and $\hat{u}(t;\boldsymbol{\theta})$ at different training epochs. Solutions after convergence are shown by solid black lines.}
    \label{fig:overview}
\end{figure}

A necessary condition for optimal control is provided by Pontryagin's maximum principle (PMP)~\cite{pontryagin1987mathematical} while the Hamilton--Jacobi--Bellman (HJB) equation provides a necessary and sufficient condition for optimality~\cite{zhou1990maximum,bellman2015applied}. To solve non-linear optimal control problems, one typically resorts to indirect and direct numerical methods. Examples of indirect optimal control solvers include different kinds of shooting methods~\cite{oberle2001bndsco} that use PMP and a control Hamiltonian to construct a system of equations describing the evolution of state and adjoint variables. In addition to indirect methods, one may also parameterize state and control functions and directly solve the resulting optimization problem. Possible function parameterizations include piecewise constant functions and other suitable basis functions~\cite{bock1984multiple}. Over the past two decades, pseudospectral methods have emerged as an effective approach for solving non-linear optimal control problems~\cite{gong2006pseudospectral} with applications in aerospace engineering~\cite{kang2007pseudospectral}. However, it has been shown that certain pseudospectral methods are not able to solve standard benchmark control problems~\cite{fahroo2008advances}. In this work, we represent time-dependent control signals by artificial neural networks (ANNs)~\cite{kang2022feedforward}. To parameterize and learn control functions, we use neural ordinary differential equations (neural ODEs)~\cite{chen2018neural,cuchiero2020deep,asikis2022neural,bottcher2022ai}. A schematic of the application of neural ODE control (NODEC) to a networked dynamical systems is shown in Fig.~\ref{fig:overview}.

Control methods that are based on ANNs have been applied to both discrete and continuous time dynamical systems~\cite{lewis2020neural}. To keep calculations associated with gradient updates tractable, applications of ANNs in control have often focused on shallow architectures and linear dynamics. For high-dimensional and non-linear dynamics with intractable gradient updates, ``identifier'' ANNs are useful to learn and replace the dynamical system underlying a given control task~\cite{lewis2020neural}. Recent advances in neural ODEs~\cite{chen2018neural} and automatic differentiation~\cite{baydin2018automatic} allow for the direct application of deep neural network architectures to high-dimensional non-linear models~\cite{asikis2022neural}, thereby avoiding identifier networks~\cite{lewis2020neural} and limitations associated with shallow ANN structures. Other popular uses of ANNs for control are found in the fields of deep reinforcement learning~\cite{PhysRevFluids.4.093902} and neural HJB methods~\cite{abu2005nearly}. Deep reinforcement learning is often applied to model-free control tasks where the model dynamics are unknown or non-differentiable~\cite{mizutani2004two}. Neural HJB methods have been applied to state-feedback control and rely on the existence of smooth value functions~\cite{abu2005nearly}. Instead of explicitly minimizing control energy or other loss functionals, neural ODE controllers have been shown to exhibit implicitly regularization properties~\cite{bottcher2022ai,asikis2022neural}. In this work, we study the effect of learning protocols and hyperparameters such as network structure and optimizers on the ability of NODEC to implicitly learn near-optimal control solutions by minimizing the distance between reached and target state.

The remainder of this paper is organized as follows. Section~\ref{sec:backprop} summarizes the basic concepts associated with neural ODE control (NODEC) and discusses two different backpropagation protocols for learning control functions: (i) truncated backpropagation through time (TBPTT) and (ii) backpropagation through time (BPTT)~\cite{williams1990efficient,werbos1990backpropagation,feldkamp1993neural}. In Section~\ref{sec:init_structure}, we discuss how parameter initialization and activation functions affect a neural network's ability to implicitly learn OC solutions. Invoking tools from dynamical systems theory, we discuss how different neural network structures affect the learnability of a constant control solution. We will show that single neurons are able to learn OC solutions only for certain initial weights and biases. Numerical results presented in Sec.~\ref{sec:nn_depth_width} indicate that as the depth of the employed neural network increases, initial conditions have a smaller impact on the ability of NODEC to learn a near-optimal control solution. In Section~\ref{sec:impl_regularization}, we analyze how gradients in the neural network parameters induce gradient updates in both the control function and control energy. By applying NODEC to a control problem with a time-dependent OC solution, we find that adaptive gradient methods such as Adam~\cite{kingma2014adam} are able to approximate OC well while steepest descent (SD) fails to do so. We complement this part of our analysis with one and two-dimensional loss projections that are useful to geometrically interpret the tradeoff between (i) a small distance between reached and target state and (ii) a small control energy. Section~\ref{sec:discussion} concludes our paper.

Our source codes are publicly available at \cite{gitlab}.
\section{Backpropagation protocols}
\label{sec:backprop}
The basic steps underlying NODEC-based solutions of optimal control problems \eqref{eq:dyn_sys} are summarized below.
\begin{enumerate}
    \item Parameterize the control input $\hat{\mathbf{u}}(t;\boldsymbol{\theta})$ using a sufficiently wide and deep neural network. Here, $\boldsymbol{\theta}\in\mathbb{R}^N$ denotes the neural-network parameters.
    \item Solve the dynamical system \eqref{eq:dyn_sys} numerically for a given initial condition ${\bf x}(0)={\bf x}_0$.
    \item Calculate the difference $L(\mathbf{x}(T),\mathbf{x}^*)$ between reached state ${\bf x}(T)$ and target state ${\bf x}^*$ and backpropagate gradients to update neural-network weights.
    \item Iterate steps 1--3 until convergence is reached (\ie, until $L(\mathbf{x}(T),\mathbf{x}^*)$ is smaller than a certain threshold).
\end{enumerate}
We will discuss in the following sections that control solutions with a small control energy $E_T[{\bf u}]$ can be obtained without explicitly including $E_T[{\bf u}]$ in the loss function. Even if the control energy is included directly in the loss function, one may still have to search for a Lagrange multiplier $\mu$ [see Eq.~\eqref{eq:oc_loss}] that is associated with a desired tradeoff between control energy and distance from the target state. We provide a corresponding example in Appendix~\ref{app:moving_particle}. Identifying the optimal value of a Lagrange multiplier~\cite{ito2008lagrange} or Lagrange costate might prove challenging or intractable in practice~\cite{thorne1996approximate}.
Implicit energy regularization properties that have been reported for NODEC~\cite{asikis2022neural,bottcher2022ai} may therefore be useful for the design of effective control methods.

To learn control functions that are suitable to solve a certain control problem minimizing the MSE between reached and target state, there exist two main classes of backpropagation protocols: (i) truncated backpropagation and approximating a desired control function within a certain subinterval of $[0,T]$ and (ii) learning the control input over the whole interval $[0,T]$ using backpropagation through time (BPTT). The first option is also referred to as truncated BPTT (TBPTT).

In the following two subsections, we will discuss the differences between BPTT and TBPTT with respect to learning control functions. We will also provide an example to compare the performance of BPTT and TBPTT.
\subsection{Truncated backpropagation through time}
For notational brevity, we will consider a TBPTT protocol for a one-dimensional flow $\dot{x}=f(x(t),\hat{u}(t;\boldsymbol{\theta}),t)$ where neural-network parameters $\boldsymbol{\theta}$ are being updated for a time $t'$ in the loss function $L=1/2(x(T)-x^*)^2$. Gradients are calculated through the loss function if the underlying dynamical system time $t$ is equal to $t'$. Generalizations for multiple evaluation times and higher-dimensional flows follow directly from the following equations. The TBPTT gradient of $L$ w.r.t.\ $\boldsymbol{\theta}$ is
\begin{align}
\begin{split}
\nabla_{\boldsymbol{\theta}}^{(t')} L &= \nabla_{\boldsymbol{\theta}}^{(t')} \frac{1}{2}\left(x(T)-x^*\right)^2 = \nabla_{\boldsymbol{\theta}}^{(t')}\frac{1}{2} \left(\int_0^{T} f(x(t),\hat{u}(t;\boldsymbol{\theta}),t)\,\mathrm{d}t+x_0-x^*\right)^2\\
&= \nabla_{\boldsymbol{\theta}}^{(t')}\frac{1}{2} \left(\int_0^{t'-\epsilon/2} f(x(t),\hat{u}(t;\boldsymbol{\theta}),t)\,\mathrm{d}t+\int_{t'+\epsilon/2}^{T} f(x(t),\hat{u}(t;\boldsymbol{\theta}),t)\,\mathrm{d}t\right.\\
& +\epsilon f(x(t'),\hat{u}(t',\boldsymbol{\theta}),t')+x_0-x^*\Bigg)^2\\
&= \epsilon \left.\frac{\mathrm{d} f}{\mathrm{d} \hat{u}} \mathcal{J}_{\hat{u}}^\top\right|_{t=t'} \left(\int_0^T f(x(t),\hat{u}(t;\boldsymbol{\theta}),t)\,\mathrm{d}t+x_0-x^*\right)\\
&=\epsilon \left.\frac{\mathrm{d} f}{\mathrm{d} \hat{u}} \mathcal{J}_{\hat{u}}^\top\right|_{t=t'} \frac{\partial L}{\partial x(T)}\\
&=\epsilon\left.\left( \frac{\partial x}{\partial \hat{u}}\frac{\partial f}{\partial x}+\frac{\partial f}{\partial \hat{u}} \right)\mathcal{J}_{\hat{u}}^\top\right|_{t=t'} \frac{\partial L}{\partial x(T)}\,,
\end{split}
\label{eq:TBPTT}
\end{align}
where $\epsilon$ is a small positive number that determines the width of the TBPTT interval. Note that $\epsilon$ can be absorbed in the learning rate associated with gradient-descent updates of $\boldsymbol{\theta}$. The Jacobian of $\hat{u}(t;\boldsymbol{\theta})$ w.r.t.\ $\boldsymbol{\theta}$ is
\begin{equation}
\mathcal{J}_{\hat{u}}=(\partial \hat{u}/\partial \theta_1,\dots,\partial \hat{u}/\partial \theta_N)\,.
\end{equation}
In the described TBPTT protocol, we used that
\begin{equation}
\nabla_{\boldsymbol{\theta}}^{(t')}f(x(t),\hat{u}(t;\boldsymbol{\theta}),t)\coloneqq
\begin{cases}
\frac{\mathrm{d} f}{\mathrm{d} \hat{u}} \mathcal{J}_{\hat{u}}^\top\quad&t= t'\\
0\quad&\text{otherwise}
\end{cases}\,.
\end{equation}
\subsection{Backpropagation through time}
Without truncating the BPTT protocol, the gradient of $L$ w.r.t.~$\boldsymbol{\theta}$ is
\begin{align}
\begin{split}
\nabla_{\boldsymbol{\theta}} L &=  \nabla_{\boldsymbol{\theta}} \frac{1}{2}\left(x(T)-x^*\right)^2 = \nabla_{\boldsymbol{\theta}}\frac{1}{2} \left(\int_0^T f(x(t),\hat{u}(t;\boldsymbol{\theta}),t)\,\mathrm{d}t+x_0-x^*\right)^2\\
&=\int_0^T \frac{\mathrm{d} f}{\mathrm{d} \hat{u}} \mathcal{J}_{\hat{u}}^\top \, \mathrm{d}t\left(\int_0^T f(x(t),\hat{u}(t;\boldsymbol{\theta}),t)\,\mathrm{d}t+x_0-x^*\right)\\
&=\int_0^T \left( \frac{\partial x}{\partial \hat{u}}\frac{\partial f}{\partial x}+\frac{\partial f}{\partial \hat{u}} \right) \mathcal{J}_{\hat{u}}^\top \, \mathrm{d}t\left(\int_0^T f(x(t),\hat{u}(t;\boldsymbol{\theta}),t)\,\mathrm{d}t+x_0-x^*\right)\\
&=I\left[D_{\hat{u}}x(T), \mathcal{J}_{\hat{u}}^\top\right]\frac{\partial L}{\partial x(T)}\,,
\end{split}
\label{eq:BPTT}
\end{align}
where the notation $I\left[D_{\hat{u}}x(T), \mathcal{J}_{\hat{u}}^\top\right]$ is used to indicate that the integration associated with the derivative $D_{\hat{u}}x(T)\coloneqq\mathrm{d} x(T)/\mathrm{d} \hat{u}$ has to be applied to $\mathcal{J}_{\hat{u}}^\top$. Note that $\partial L/\partial x(T)$ is not time-dependent, while the derivative of the loss function $L$ w.r.t.\ $\hat{u}$ is an operator that acts on the time-dependent quantity $\mathcal{J}_{\hat{u}}^\top$.

We will now compare BPTT and TBPTT protocols for controlling the two-dimensional linear flow
\begin{figure}
    \centering
    \includegraphics[width=\textwidth]{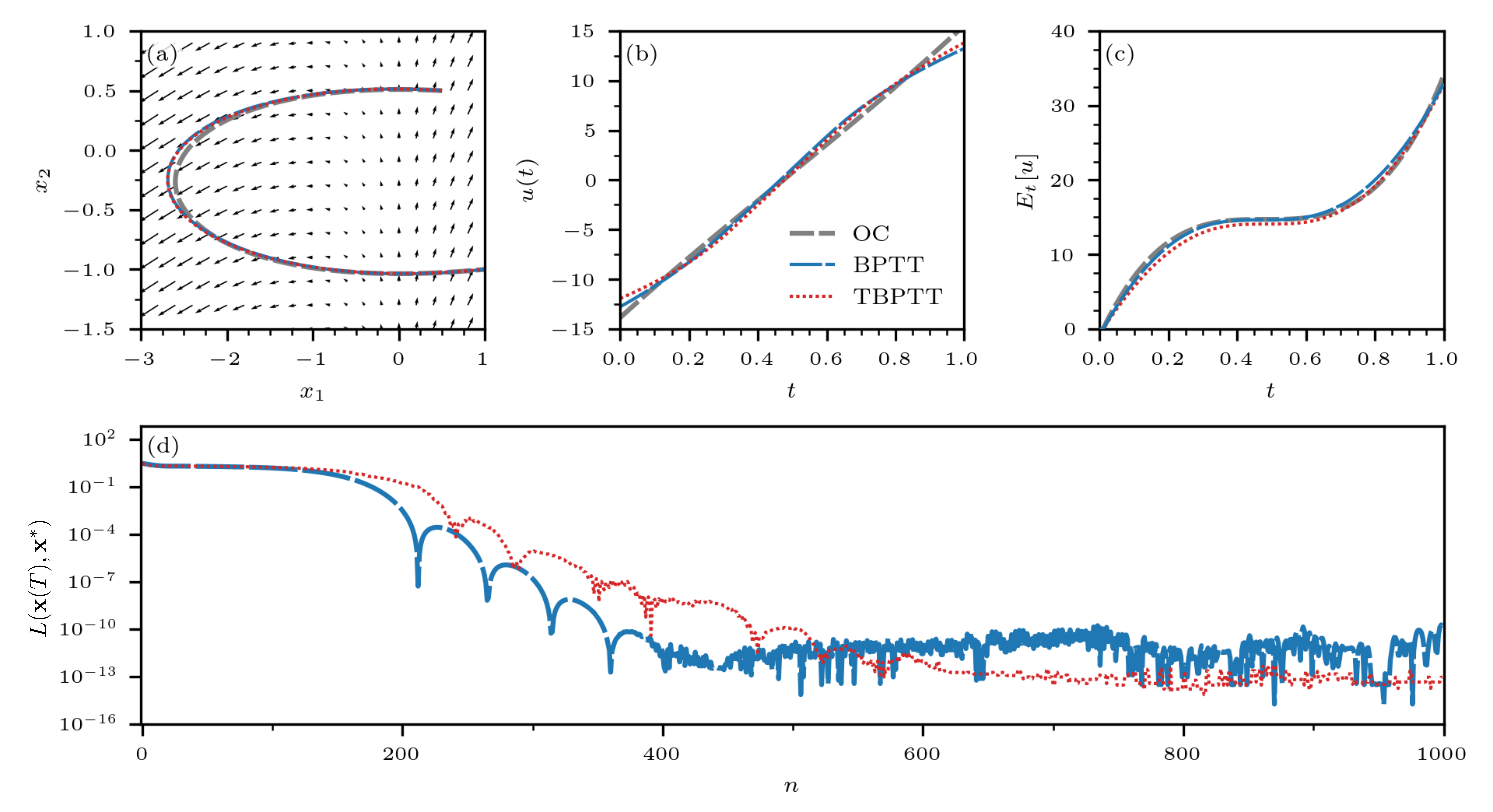}
    \caption{Learning control functions with different backpropagation protocols. (a) The evolution of $\mathbf{x}(t)$ of the solution of Eq.~\eqref{eq:2d_flow} in the $(x_1,x_2)$ plane. (b,c) The evolution of the corresponding control function and control energy. The OC solution is indicated by a dashed grey line. Black disks and red triangles show the solutions associated with BPTT and TBPTT protocols, respectively. The underlying neural network has 2 hidden ELU layers and 14 neurons per hidden layer. We trained the neural network for 1,000 epochs using Adam. The learning rates are $\eta=3\times10^{-3}$ (BPTT) and $\eta=5\times 10^{-3}$ (TBPTT). (d) The loss $L(\mathbf{x}(T),\mathbf{x}^*)$ as a function of training epochs $n$. Both backpropagation protocols can achieve low loss values after a few hundred epochs of training.
    }
    \label{fig:btt_tbtt}
\end{figure}
\begin{equation}
\dot{\mathbf{x}}=A \mathbf{x}+B\mathbf{u}\,,\quad {\bf x}(0)=(0.5,0.5)^\top\,,\quad {\bf x}^*=(1,-1)^\top\,,
\label{eq:2d_flow}
\end{equation}
with
\begin{equation}
A = \begin{pmatrix}
1 & 0\\
1 & 0
\end{pmatrix}\quad \text{and}\quad 
B = \begin{pmatrix}
1\\
0
\end{pmatrix}\,.
\label{eq:linear_system}
\end{equation}

The optimal control $\mathbf{u}^*(t)$ associated with solving Eq.~\eqref{eq:2d_flow} and minimizing $E_T[\mathbf{u}]$ [see Eq.~\eqref{eq:control_energy}] has been derived in \cite{yan2012controlling}. We compare the optimal solution with the solutions obtained using (T)BPTT and the MSE loss
\begin{equation}
L=\frac{1}{2}\left\|\mathbf{x}(T)-\mathbf{x}^*\right\|_2^2\,.
\end{equation}
Figure~\ref{fig:btt_tbtt}(a) shows the evolution of $\mathbf{x}(t)$ in the $(x_1,x_2)$ plane. Both backpropagation protocols, BPTT and TBPTT, produce solutions after 2,000 epochs of training that are similar to the OC solution. This similarity is also reflected in the evolution of the control signal and control energy [see Fig.~\ref{fig:btt_tbtt}(b,c)]. To solve Eq.~\eqref{eq:2d_flow} numerically, we used a forward Euler method with step size 0.01. In the TBTT protocol, we use a different gradient evaluation time $t'$ in each backpropagation step. During experiments we observe that convergence of both methods is sensitive to choice of learning rate and optimizer.
Figure~\ref{fig:btt_tbtt}(d) shows that BPTT converges faster with a smaller learning rate, but at later stages in training it becomes less stable. On the contrary, TBPTT, converges slower with a higher learning rate, but it is more stable after convergence. Under the employed parametrization, both methods are capable to converge fast to low error values. In terms of computation time, on the same hardware and according to \texttt{tqdm} library measurements, BPTT performs around 65 training epochs per second, whereas TBPTT achieves approximately 125 epochs per second. The difference in total computation time may be attributed to the fewer gradient evaluations required by TBPTT.
\section{Parameter initialization and activation functions}
\label{sec:init_structure}
As shown in Sec.~\ref{sec:backprop} and in previous work~\cite{bottcher2022ai,asikis2022neural}, NODEC is able to learn near-optimal control solutions without explicitly minimizing $E_T[{\bf u}]$. To provide insights into the learning dynamics underlying NODEC, we study two control problems associated with a one-dimensional linear flow. The first dynamical system has a constant OC solution while that of the second dynamical system is time-dependent. In this section, we will show that shallow architectures require one to tune initial weights and biases to learn OC solutions without accounting for a control energy term in the loss function. The next section then provides numerical evidence that deeper architectures are less prone to such weight and bias initialization effects.
\subsection{Constant control}
\label{sec:constant_control}
We first consider a state-independent, one-dimensional flow
\begin{equation}
\dot{x}=u\,,\quad {x}(0)={x}_0\,,\quad {x}(T)={x}^*\,.
\label{eq:dyn_sys_simple}
\end{equation}
The control Hamiltonian associated with Eqs.~\eqref{eq:dyn_sys_simple} and \eqref{eq:control_energy} is
\begin{equation}
H = \lambda u(t)+\frac{1}{2} u(t)^2\,.
\end{equation}
Using the control Hamiltonian, we obtain the optimal control $u^*(t)$ from PMP according to
\begin{align}
\begin{split}
0&=\left.\frac{\partial H}{\partial u}\right|_{u=u^*}=\lambda + u^*(t)\,,\\
\dot{\lambda}&=-\frac{\partial H}{\partial x}=0\,.
\end{split}
\end{align}
Integrating Eq.~\eqref{eq:dyn_sys_simple} with $u^*(t)=-\lambda$ yields
\begin{equation}
x(t)=x_0-\int_0^t \lambda\,\mathrm{d}t'=x_0+\frac{t(x^*-x_0)}{T}\,,
\end{equation}
where we used that the reached state $x(T)$ is equal to the target state $x^*$. The optimal control is $u^*=(x^*-x_0)/T$.

\begin{figure}
    \centering
    \includegraphics[width=0.8\textwidth]{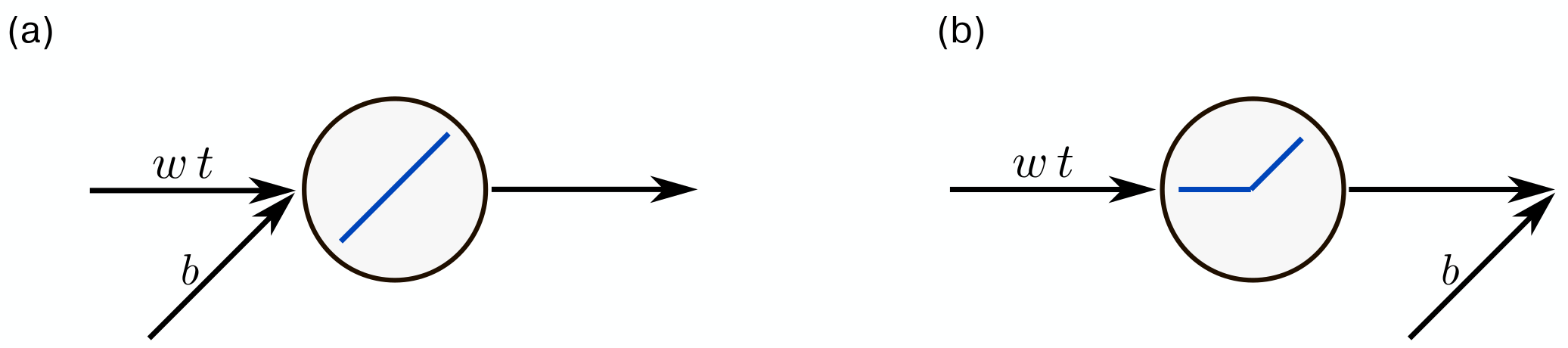}
    \caption{Examples of single-neuron structures. (a) The input is the time $t$ and the output is $w t+b$. (b) The input is the time $t$ and the output is $\max(0,w t)+b$.}
    \label{fig:schematic_1}
\end{figure}

A simple neural network that consists of a single ReLU and uses the time $t$ as an input is, in principle, able to represent the constant optimal control $u^*$. For a neural-network generated control signal $\hat{u}(t;\boldsymbol{\theta})={\rm ReLU}(w\,t+b)=\max(0,w\,t+b)$, where $\boldsymbol{\theta}=(w,b)^\top$, we have $\hat{u}(t;\boldsymbol{\theta})=u^*(t)$ if $w^*=0$ and $b^*=(x^*-x_0)/T$.

Near-optimal control solutions have been obtained representing control functions by neural ODEs and optimizing
\begin{equation}
L=\frac{1}{2}(x(T)-x^*)^2\,,
\label{eq:nn_loss}
\end{equation}
without explicitly accounting for the integrated cost or control energy \eqref{eq:control_energy}~\cite{bottcher2022ai}. Under what conditions (\eg, initial weights and biases, optimization algorithm, and neural-network structure) can a neural network learn a control function $\hat{u}(t;\boldsymbol{\theta})$ that is close to $u^*(t)$ by optimizing Eq.~\eqref{eq:nn_loss}? As a starting point for addressing this question, we will focus on analytically tractable learning algorithms and neural network structures. In the following sections, we will then use these gained insights to draw conclusions for control problems that involve higher-dimensional neural network structures and generalized optimization algorithms.

\begin{figure}
    \centering
    \includegraphics[width=\textwidth]{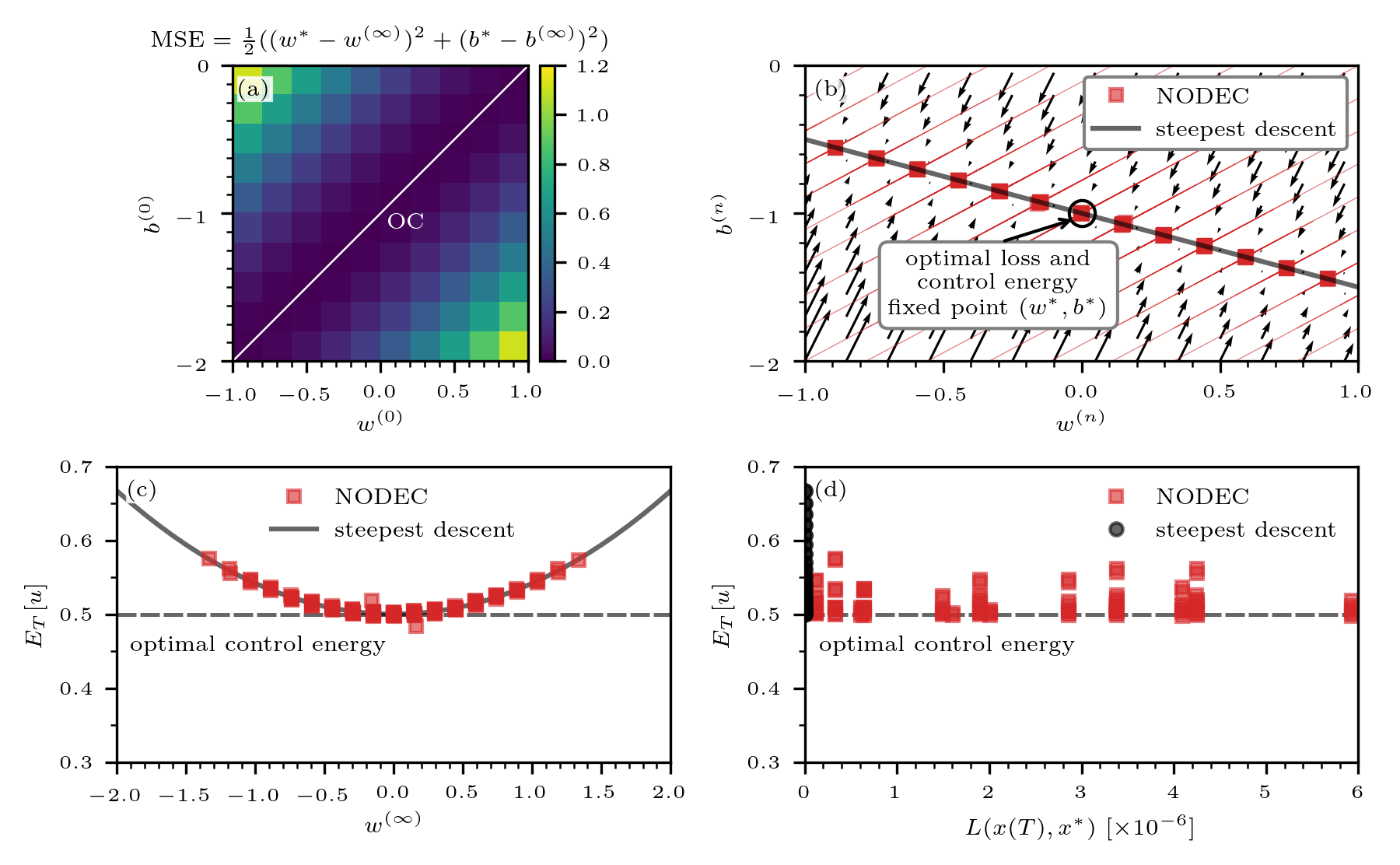}
    \caption{Approximating constant control with a linear activation function. (a) The mean squared error (MSE) associated with the deviation of the learned weight and bias, $(w^{(\infty)},b^{(\infty)})$, from the optimal control weight and bias, $(w^*,b^*)=(0,-1)$, for different initial weights $w^{(0)}$ and biases $b^{(0)}$. The solid white line indicates the location of initial weights and biases for which gradient-descent learning converges to optimal control. (b) Convergence of weight and bias values towards the attractor $w^{(\infty)}= -2(1 + b^{(\infty)})$. Solid red lines and filled red squares indicate learning trajectories and learned parameters $(w^{(\infty)},b^{(\infty)})$, respectively. The solid black line represents fixed points of gradient-descent learning [\ie, $w^{(\infty)}= -2(1 + b^{(\infty)})$ as defined in Eq.~\eqref{eq:fixed_points_1}]. The fixed point $(w^*,b^*)=(0,-1)$ corresponding to optimal control is marked by a black circle. (c) The control energy $E_T[u]$ [see Eq.~\eqref{eq:control_energy}] as a function of learned neural-network weights $w^{(\infty)}$. Energies associated with simulation results and gradient-descent fixed points are indicated by filled red disks and a solid black line, respectively. The dashed grey line marks the energy associated with optimal control. (d) Control energy as a function of the loss function \eqref{eq:simple_loss}. In all simulations, we set $T=1$, $x_0=0$, and $x^*=-1$. We trained the controller with Adam and used a learning rate $\eta=0.1$.}
    \label{fig:simple_dynamics_1}
\end{figure}
\subsubsection{Single-neuron structure with linear activation}
In the following example, we use simple neural network that consists of a linear activation and a weight and bias parameter [see Fig.~\ref{fig:schematic_1}(a)]. The corresponding loss function is
\begin{align}
\begin{split}
L(x(T),x^*)&=\frac{1}{2}(x(T)-x^*)^2=\frac{1}{2}\left(\int_0^T (w t+b) \,\mathrm{d}t+x_0-x^*\right)^2\\
&=\frac{1}{2}\left(\frac{1}{2} w T^2+b T+x_0-x^*\right)^2\,.
\end{split}
\label{eq:simple_loss}
\end{align}
Using steepest descent learning and BPTT, weights and biases are iteratively updated according to
\begin{align}
\begin{split}
w^{(n+1)}&=w^{(n)}-\eta \frac{\partial L}{\partial w^{(n)}}\,,\\
b^{(n+1)}&=b^{(n)}-\eta \frac{\partial L}{\partial b^{(n)}}\,,
\end{split}    
\label{eq:gradients}
\end{align}
where $w^{(0)}$ and $b^{(0)}$ denote initial weight and bias, respectively. The quantity $\eta$ denotes the learning rate. Using the loss function \eqref{eq:simple_loss}, the gradients in Eq.~\eqref{eq:gradients} are
\begin{align}
\begin{split}
\frac{\partial L}{\partial w^{(n)}}&=\frac{1}{2}T^2\left(\frac{1}{2}w^{(n)} T^2+b^{(n)} T+x_0-x^*\right)\,,\\
\frac{\partial L}{\partial b^{(n)}}&=T\left(\frac{1}{2}w^{(n)} T^2+b^{(n)}T+x_0-x^*\right)\,.
\end{split}
\label{eq:gradient_descent_1}
\end{align}
The fixed points $(w^{(\infty)},b^{(\infty)})$ of the iteration \eqref{eq:gradient_descent_1} satisfy
\begin{equation}
w^{(\infty)}= -\frac{2}{T^2}(b^{(\infty)} T+x_0-x^*)\,.
\label{eq:fixed_points_1}
\end{equation}
For $T=1$, $x_0=0$, and $x^*=-1$, the fixed points are given by $w^{(\infty)}= -2(1 + b^{(\infty)})$. Recall that, in this example, the weight and bias of optimal control are $w^*=0$ and $b^*=-1$. This fixed point can only be reached for specific initializations as shown in Fig.~\ref{fig:simple_dynamics_1}(a,b). Other fixed points are associated with certain tradeoffs between small control energies
\begin{equation}
E_T[u]=\frac{1}{2}\int_0^T \left[ b^{(\infty)}-2(1+b^{(\infty)})\,t\right]^2\,\mathrm{d}t=\frac{1}{6} [4 + b^{(\infty)} (2 + b^{(\infty)})]\,.
\end{equation}
and small losses $L(x(T),x^*)$ [see Fig.~\ref{fig:simple_dynamics_1}(c,d)]. 
\subsubsection{Single-neuron structure with ReLU activation}
We now use a slightly different neural-network structure that consists of a single ReLU [see Fig.~\ref{fig:schematic_1}(b)]. The loss function in this example is
\begin{equation}
\begin{split}
L(x(T),x^*)&=\frac{1}{2}(x(T)-x^*)^2=\frac{1}{2}\left(T+\int_0^T (\max(0,w\,t)+b)\,\mathrm{d}t+x_0-x^*\right)^2\\
&=
\begin{cases}
\frac{1}{2}\left(\frac{1}{2}w\,T^2+bT+x_0-x^*\right)^2\quad&\text{if $w\geq 0$}\\
\frac{1}{2}\left(bT+x_0-x^*\right)^2\quad&\text{otherwise}
\end{cases}
\,.
\end{split}
\label{eq:simple_loss_2}
\end{equation}
If the weights are positive, the gradient-descent equations are equivalent to Eq.~\eqref{eq:gradient_descent_1}. Otherwise, we have
\begin{equation}
\begin{split}
w^{(n+1)}&=w^{(n)}\,,\\
b^{(n+1)}&=b^{(n)}-\eta T(b^{(n)}T+x_0-x^*)\,.
\end{split}
\label{eq:gradient_descent_2}
\end{equation}
\begin{figure}
    \centering
    \includegraphics[width=\textwidth]{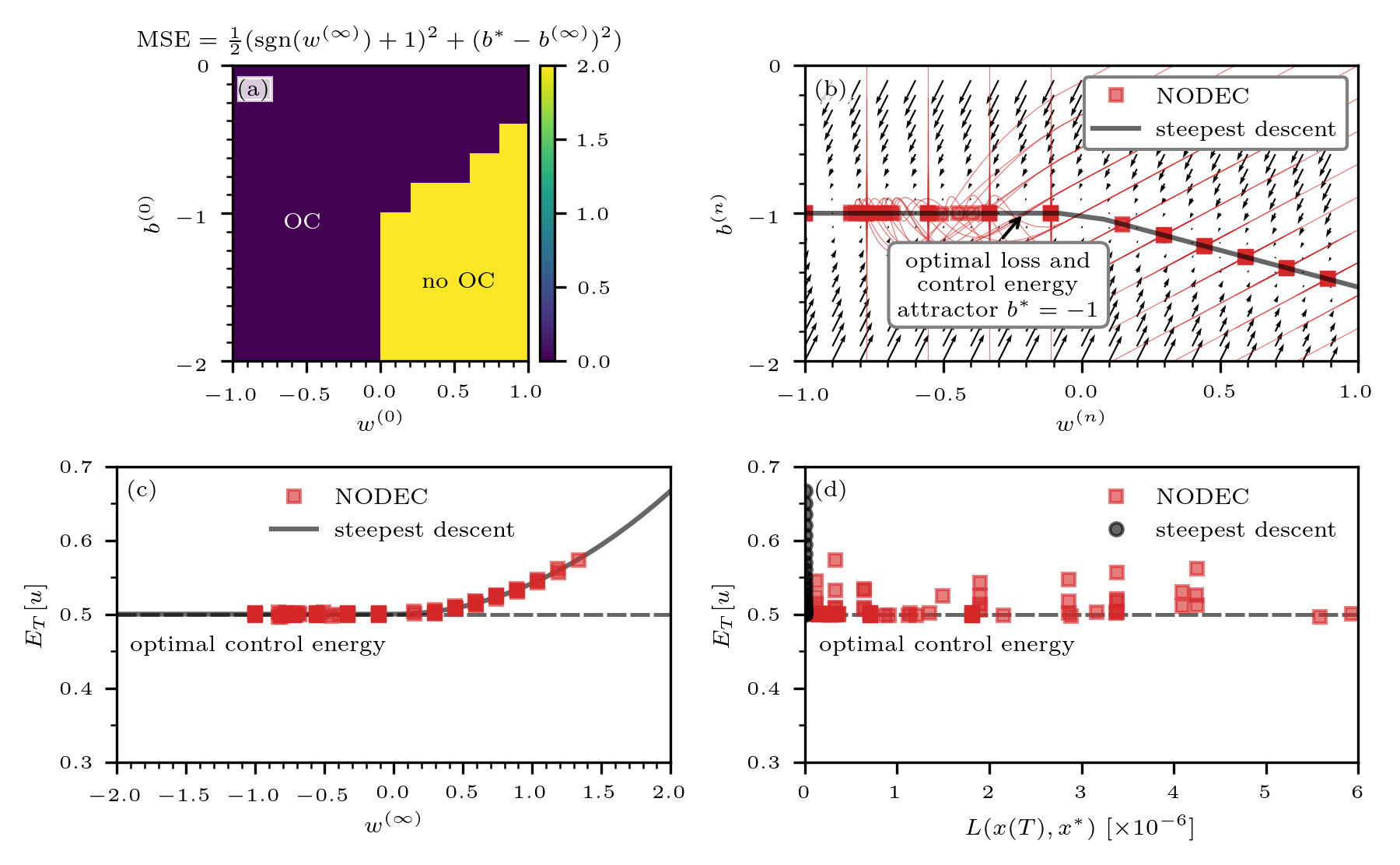}
    \caption{Approximating constant control with a ReLU´ activation function. (a) The mean squared error (MSE) associated with the deviation of the learned weight and bias, $(w^{(\infty)},b^{(\infty)})$, from the optimal control weight and bias, $w^*\leq 0,b^*=-1$, for different initial weights $w^{(0)}$ and biases $b^{(0)}$. The deep purple region indicates initial weights and biases for which the neural-network learns optimal control. Optimal control cannot be learned in the yellow region.  (b) Convergence of weight and bias values towards the attractor $w^{(\infty)}= -2(1 + b^{(\infty)})$. Solid red lines and filled red squares indicate learning trajectories and learned parameters $(w^{(\infty)},b^{(\infty)})$, respectively. The solid black line represents the attractor of gradient-descent learning [\ie, $b^*=-1$ if $w^{(\infty)}\leq 0$ and $w^{(\infty)}= -2(1 + b^{(\infty)})$ if $w^{(\infty)}>0$]. The attractor $w^*\leq 0,b^*=-1$ corresponds to optimal control. (c) The control energy $E_T[u]$ [see Eq.~\eqref{eq:control_energy}] as a function of learned neural-network weights $w^{(\infty)}$. Energies associated with simulation results and gradient-descent fixed points are indicated by filled red disks and the solid black line, respectively. The dashed grey line marks the energy associated with optimal control. (d) Control energy as a function of the loss function \eqref{eq:simple_loss}. In all simulations, we set $T=1$, $x_0=0$, and $x^*=-1$. We trained the controller with Adam and used a learning rate $\eta=0.1$.}
    \label{fig:simple_dynamics_2}
\end{figure}

If $w^{(n)}<0$ for some $n$, the fixed point of the gradient descent \eqref{eq:gradient_descent_2} is equivalent to the optimal control $b^{(\infty)}=b^{*}=\frac{x^*-x_0}{T}$. Figure~\ref{fig:simple_dynamics_2}(a,b) shows that the optimal-control domain of Eq.~\eqref{eq:simple_loss_2} is substantially larger than that resulting from a gradient descent in Eq.~\eqref{eq:simple_loss}. The control energy approaches that of optimal control for $w^{(\infty)}\leq 0$, while larger values of $w^{(\infty)}$ are associated with larger control energies [see Fig.~\ref{fig:simple_dynamics_2}(c)]. As in the previous example, one can observe that tradeoffs between small control energies and small losses are possible.

Equation \eqref{eq:gradient_descent_2} also shows that optimal control of Eq.~\eqref{eq:dyn_sys_simple} is a global attractor if one would just learn a bias term, reflecting the fact that the optimal control is given by a constant.
\subsection{Time-dependent control}
\label{sec:time_dep_control}
\begin{figure}
    \centering
    \includegraphics[width=\textwidth]{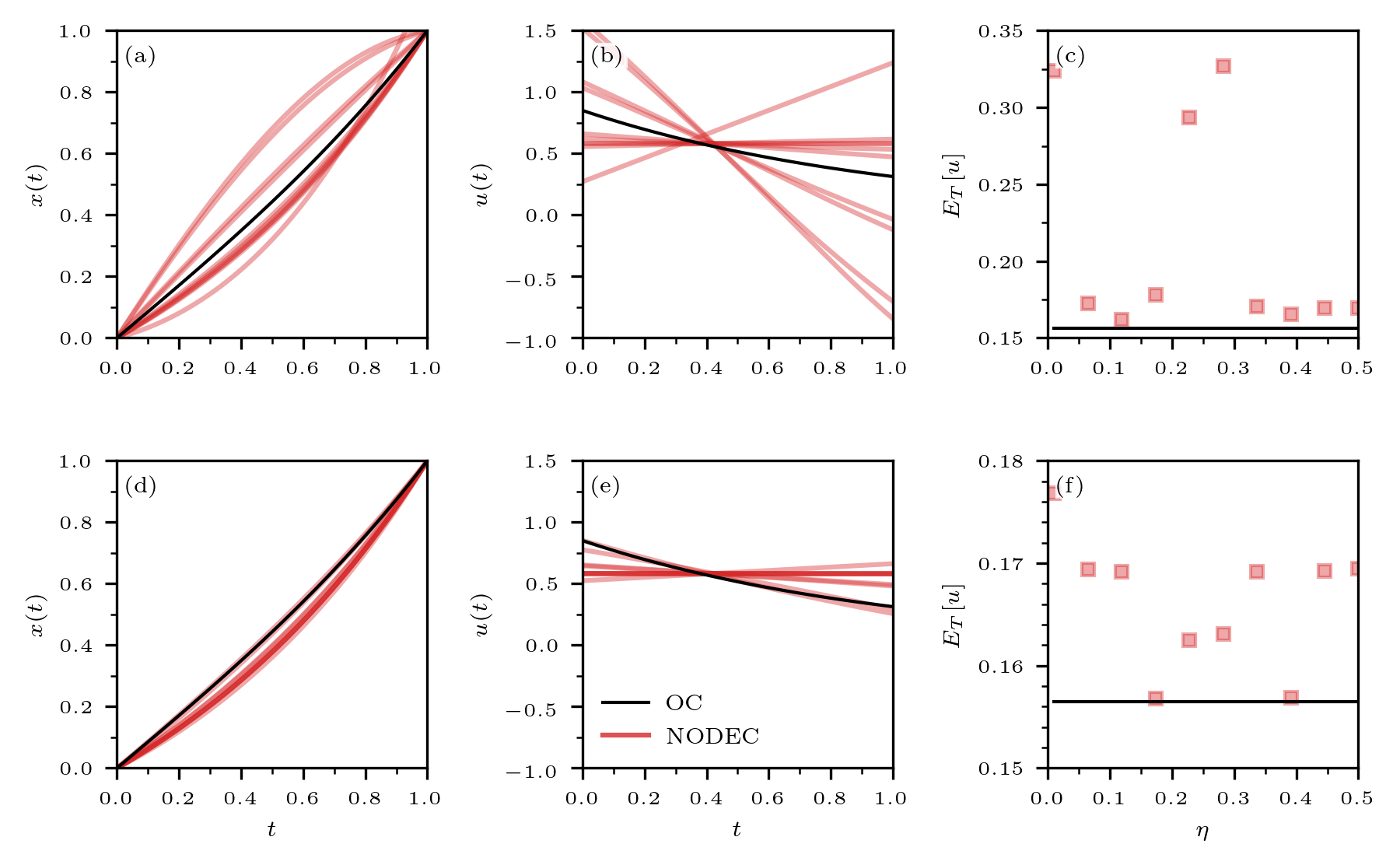}
    \caption{Approximating time-dependent control. We use NODEC to control Eq.~\eqref{eq:dyn_sys_simple_time} for $x_0=0$ and $T=a=b=x^*=1$. The neural network that we use in this example consists of 2 hidden layers with 6 ELU neurons each. (a--c) Initial weights and biases are set to $1$. (d--f) Initial weights and biases are set to $0.1$. Different solid red lines are associated with different learning rates. Solid black lines indicate OC solutions \eqref{eq:time_dependent_control}--\eqref{eq:control_energy_time_dependent}. We trained NODEC for $10^3$ epochs using Adam.}
    \label{fig:time_dependent_control}
\end{figure}
To study the ability of NODEC to learn time-dependent control functions, we consider the linear dynamical system 
\begin{equation}
\dot{x}=ax+bu\,,\quad {x}(0)={x}_0\,,\quad {x}(T)={x}^*\,.
\label{eq:dyn_sys_simple_time}
\end{equation}
The corresponding optimal control signal that satisfies Eqs.~\eqref{eq:u_min} and \eqref{eq:dyn_sys} is
\begin{equation}
u^*(t)=\frac{a e^{-a t}}{\sinh(a T)} \left(x^*-x_0e^{a T}\right)\,,
\label{eq:time_dependent_control}
\end{equation}
Note that for $a>0$ the magnitude of $u^*(t)$ decays exponentially because larger values of $t$ are associated with larger state values of the uncontrolled dynamics $\dot{x}=a x$. The opposite holds for $a<0$. The evolution of the system state $x(t)$ under the influence of $u^*(t)$ is
\begin{equation}
x^*(t)=x_0 e^{a t}+\frac{\sinh (a t)}{\sinh (a T)} \left(x^*-x_0 e^{a T}\right)\,,
\end{equation}
and the control energy associated with the optimal control signal \eqref{eq:time_dependent_control} is
\begin{equation}
E_T[u^*]=\frac{1}{2}\int_0^T {u^*(t)}^2\,\mathrm{d}t=\frac{a(1-e^{-2 a T})(x^*-x_0 e^{a T})^2}{4 \sinh^2(a T)}\,.
\label{eq:control_energy_time_dependent}
\end{equation}

As in the prior examples, we use NODEC to learn $\hat{u}(t;\boldsymbol{\theta})$ by minimizing the MSE loss \eqref{eq:nn_loss}. We set $x_0=0$ and $T=a=b=x^*=1$, such that $x^*(t)=\sinh(t)/\sinh(1)$, $u^*(t)=e^{-t}/\sinh(1)$, and $E_T[u^*]=1/(e^2-1)\approx 0.157$. To capture the time-dependence of the control input, we use a neural network with two hidden layers and six ELU units per hidden layer. Numerical experiments with smaller architectures showed that the exponential decay of the OC solution could not be captured well. The number of timesteps in our simulations is 100. If bias and weight values are initialized to 1, NODEC is able to steer the dynamical system towards the desired target state [see Fig.~\ref{fig:time_dependent_control}(a)]. For a learning rate of about $0.12$, we observe that NODEC approximates the OC solution by a constant control with a similar control energy [see Fig.~\ref{fig:time_dependent_control}(b,c)]. The relative difference between the two control energies is about 3.5\%. Using smaller initial weights and biases can help NODEC approximate the OC solution even more closely [see Fig.~\ref{fig:time_dependent_control}(d--f)]. We find a minimum relative control energy difference of less than 0.2\%. Earlier work~\cite{bottcher2022ai} also suggested that small initial weights and biases can be helpful for learning OC solutions without explicitly regularizing control energy. 
\section{Neural network depth and width}
\label{sec:nn_depth_width}
After having discussed the effect of parameter initialization on the ability of NODEC to implicitly learn OC solutions, we will now study how different numbers of layers and neurons per layer can help improve learning performance in the context of the examples from Sec.~\ref{sec:init_structure}.
\subsection{Constant control}
\begin{figure}
    \centering
    \includegraphics[width=\textwidth]{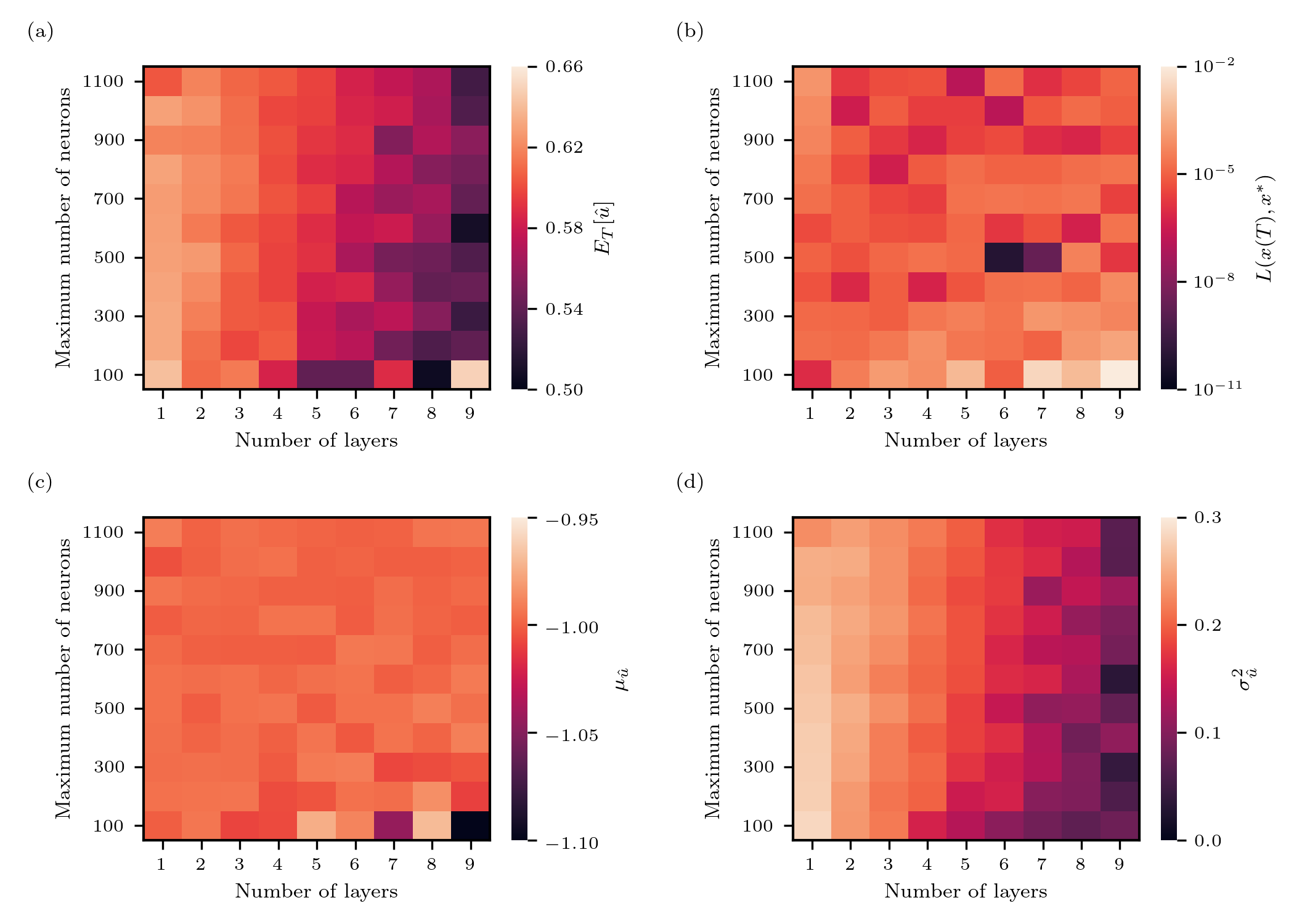}
    \caption{Effect of variations in number of layers and neurons per layer on the ability of NODEC to learn a constant control signal. The underlying architectures use hyperbolic tangent activations in all hidden layers. Bias terms are not included. Each ANN is trained for 100 epochs using Adam and a learning rate $\eta=10^{-3}$. We set the number of neurons per layer to be the greatest integer less than or equal to the maximum number of neurons (vertical axes) divided by the number of layers (horizontal axes).
    The optimal constant control is $-1$ and the optimal control energy is $1/2$. Heatmaps show the (a) control energy $E_T[\hat{u}]$, (b) loss $L(x(T),x^*)$, (c) mean control signal over time $\mu_{\hat{u}}$, and (d) control signal variance over time, $\sigma^2_{\hat{u}}$.
    }
    \label{fig:depth_vs_width_constant}
\end{figure}
We first focus on the ability of deeper ANN architectures to implicitly learn a constant control function for the example that we discussed in Sec.~\ref{sec:constant_control}. The optimal control signal is $u^*=-1$, and we use an ANN with hyperbolic tangent activations, between one to nine layers, and up to 1,100 neurons. Weights and biases are initially distributed according to $\mathcal{U}(-\sqrt{k},\sqrt{k})$, where $k$ denotes the number of input features of a certain layer. Figure~\ref{fig:depth_vs_width_constant}(a) shows that an increase in the number of layers is associated with a decrease in control energy towards the optimal value of 0.5. Even if the number of layers increases, loss values are still small and almost unaffected by the change of the ANN architecture [see Fig.~\ref{fig:depth_vs_width_constant}(b)]. Increasing the number of layers also helps in learning constant controls, as the temporal mean $\mu_{\hat{u}}$ is close to the optimal constant control for most simulated scenarios with many layers [see Fig.~\ref{fig:depth_vs_width_constant}(c)] while the corresponding variance $\sigma^2_{\hat{u}}$ gets closer to 0 [see Fig.~\ref{fig:depth_vs_width_constant}(d)]. To summarize, we observe that adding more layers is more beneficial than adding more neurons in order to approximate a constant OC function.
\subsection{Time-dependent control}
\begin{figure}
    \centering
    \includegraphics[width=\textwidth]{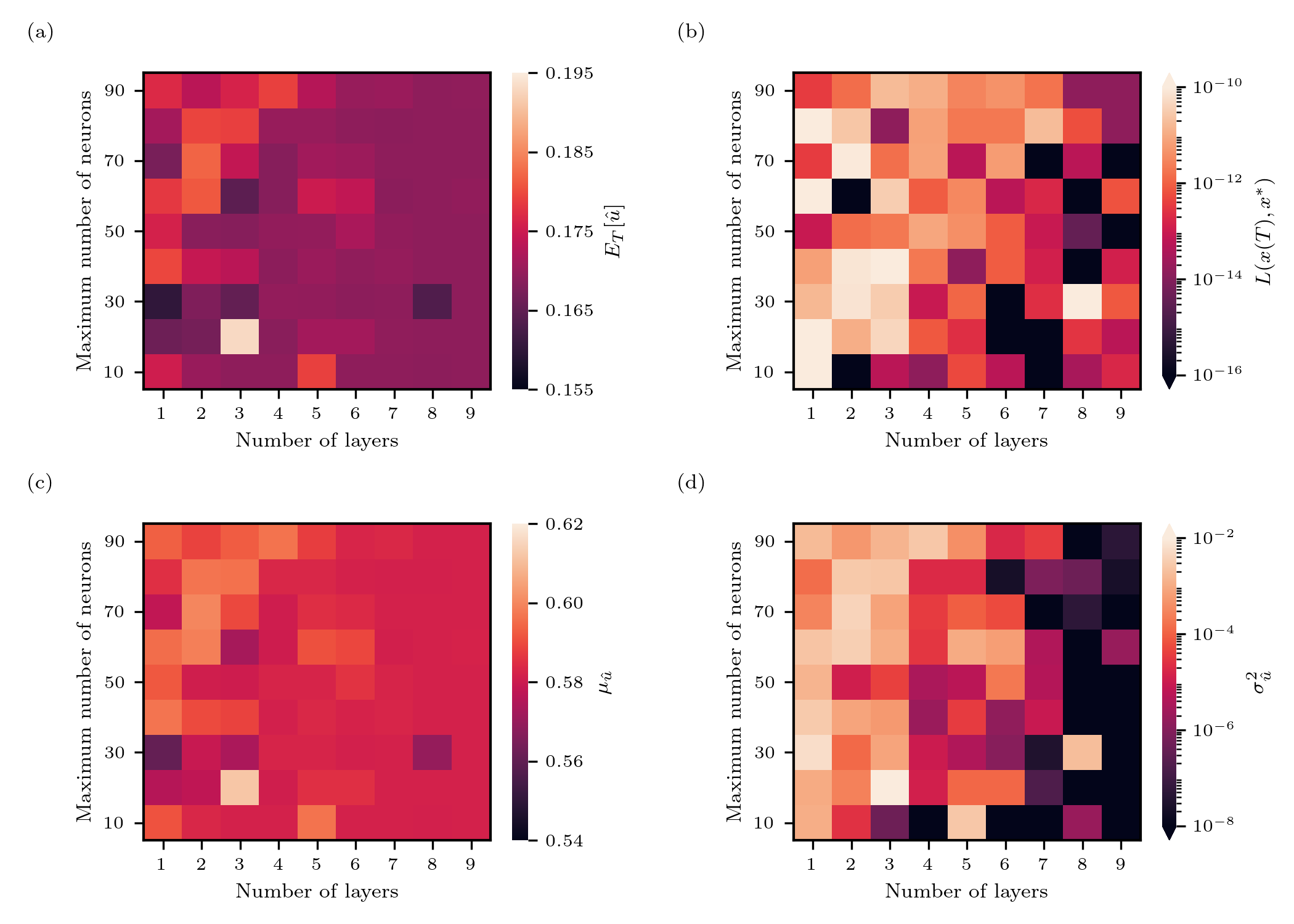}
    \caption{Effect of variations in number of layers and neurons per layer on the ability of NODEC to learn the time-dependent control signal \eqref{eq:time_dependent_control} for $x_0=0$ and $T=a=x^*=1$. The underlying architectures use bias terms and ELU activations in all hidden layers. Each ANN is trained for 500 epochs using Adam and a learning rate $\eta=3 \times 10^{-3}$. We set the number of neurons per layer to be the greatest integer less than or equal to the maximum number of neurons (vertical axes) divided by the number of layers (horizontal axes). The optimal control energy is approximately $0.157$. Heatmaps show the (a) control energy $E_T[\hat{u}]$, (b) loss $L(x(T),x^*)$, (c) mean control signal over time $\mu_{\hat{u}}$, and (d) control signal variance over time, $\sigma^2_{\hat{u}}$.
    }
    \label{fig:depth_vs_width_linear}
\end{figure}
For learning the time-dependent control signal that we studied in Sec.~\ref{sec:time_dep_control}, we use an ANN with ELU activations, between one to nine layers, and up to 90 neurons. Weights and biases are initially distributed according to $\mathcal{U}(-\sqrt{k},\sqrt{k})$, where $k$ denotes the number of input features of a certain layer. Figure~\ref{fig:depth_vs_width_linear}(a) shows that smaller control energies can be achieved on average as the number of layers increases. For one or eight layers and 30 neurons in total, we observe that the ANN controller is able to implicitly learn a solution with a small loss and a control energy that is close to that of the optimal solution. However, in general, the loss may increase with the number of layers [see Fig.~\ref{fig:depth_vs_width_linear}(b)].
Increasing the layers while keeping a low number of neurons seems to affect convergence as most ANNs seem to converge to a non-optimal constant control\textemdash the variance in the right side of Fig.~\ref{fig:depth_vs_width_linear}(d) is close to 0 (darker color). We show in Sec.~\ref{sec:const_contr_approx} that the constant OC approximation of Eq.~\eqref{eq:time_dependent_control} has a control energy of about 0.169 [see Fig.~\ref{fig:depth_vs_width_linear}]. In this example, we find that implicit regularization of control energy can be achieved by selecting appropriate hyperparameters (\eg, one or eight layers and 30 neurons in total ).

In Appendix~\ref{app:time_dependent}, we study the effect of different numbers of layers and neurons on the performance of NODEC in implicitly learning near-optimal control solutions for the time-dependent control signal associated with the two-dimensional flow discussed in Sec.~\ref{sec:backprop}.
\section{Implicit energy regularization}
\label{sec:impl_regularization}
\begin{figure}
    \centering
    \includegraphics[width=\textwidth]{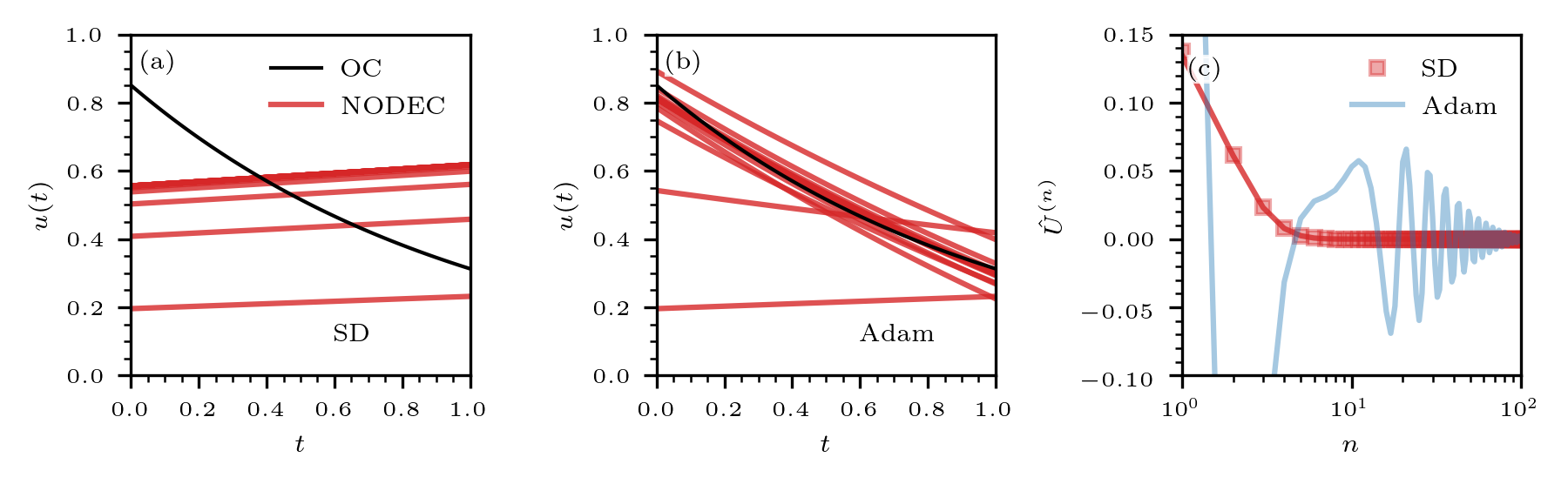}
    \caption{Learning control signals with steepest descent and Adam. We use NODEC to control Eq.~\eqref{eq:dyn_sys_simple_time} for $x_0=0$ and $T=a=b=x^*=1$. (a,b) Learned and optimal control signals are indicated by solid red and black lines, respectively [(a) steepest descent; (b) Adam]. Different solid red lines correspond to control functions $\hat{u}(t;\boldsymbol{\theta}^{(n)})$ learned after different numbers of epochs $n$. The learning rate was set to $\eta=0.15$. The neural network that we use in this example consists of 2 hidden layers with 6 ELU neurons each. Initial weights and biases are set to 0.1. Solid black lines indicate the optimal control signal \eqref{eq:time_dependent_control}. (c) The weighted total change $\Delta \hat{U}^{(n)}$ [see Eq.~\eqref{eq:delta_U_n}] as a function of epochs $n$. Red squares (steepest descent) and the blue solid line (Adam) are based on directly evaluating $\int_0^T\hat{u}(t;\boldsymbol{\theta}^{(n+1)})e^{-a t}\,\mathrm{d}t-\int_0^T\hat{u}(t;\boldsymbol{\theta}^{(n)})e^{-a t}\,\mathrm{d}t$. The solid red line is based on evaluating the right-hand side of Eq.~\eqref{eq:delta_U_n} using neural-network parameter changes $\Delta \boldsymbol{\theta}^{(n)}$ associated with steepest descent.}
    \label{fig:learning_u_GD_vs_Adam}
\end{figure}
For sufficiently small changes in the neural-network parameters $\Delta \boldsymbol{\theta}^{(n)}$ (\ie, a sufficiently small learning rate $\eta$), a steepest-descent update in $\boldsymbol{\theta}^{(n)}$ changes a control input according to
\begin{align}
\begin{split}
\hat{\mathbf{u}}(t;\boldsymbol{\theta}^{(n+1)})&=\hat{\mathbf{u}}(t;\boldsymbol{\theta}^{(n)})+\mathcal{J}_{\hat{\mathbf{u}}}\Delta \boldsymbol{\theta}^{(n)}=\hat{\mathbf{u}}(t;\boldsymbol{\theta}^{(0)})+\sum_{i=0}^n \mathcal{J}_{\hat{\mathbf{u}}^{(i)}} \Delta \boldsymbol{\theta}^{(i)}\\
&=\hat{u}(t;\boldsymbol{\theta}^{(0)})-\eta \sum_{i=0}^n \mathcal{J}_{\hat{\mathbf{u}}^{(i)}} \nabla_{\boldsymbol{\theta}^{(i)}}L\,,
\end{split}
\label{eq:induced_descent}
\end{align}
where $\Delta \boldsymbol{\theta}^{(n)}=-\eta \nabla_{\boldsymbol{\theta}^{(n)}}L$. The gradient $\nabla_{\boldsymbol{\theta}^{(n)}}L$ can be evaluated with BPTT and TBPTT protocols [see Eqs.~\eqref{eq:TBPTT} and \eqref{eq:BPTT}]. We will now study the evolution of control inputs under neural-network parameter updates.

\subsection{Induced gradient update}
For the linear one-dimensional flow \eqref{eq:dyn_sys_simple_time}, we have
\begin{equation}
x(t)=e^{a t}\left(x_0+\int_0^t b e^{-a t'}\hat{u}(t';\boldsymbol{\theta}^{(n)})\,\mathrm{d}t'\right)
\end{equation}
and
\begin{equation}
L=\frac{1}{2}\left(x(T)-x^*\right)^2=\frac{1}{2}\left[e^{a T}\left(x_0+\int_0^T b e^{-a t}\hat{u}(t;\boldsymbol{\theta}^{(n)})\,\mathrm{d}t\right)-x^*\right]^2\,.
\end{equation}
Using the BPTT notation introduced in Sec.~\ref{sec:backprop}, we find
\begin{equation}
\begin{split}
\nabla_{\boldsymbol{\theta}^{(n)}} L&=b e^{a T}\int_0^Te^{-a t}\nabla_{\theta^{(n)}} \hat{u}(t;\boldsymbol{\theta}^{(n)})\,\mathrm{d}t\left[e^{a T}\left(x_0+\int_0^T b e^{-a t}\hat{u}(t;\boldsymbol{\theta}^{(n)})\,\mathrm{d}t\right)-x^*\right]\\
&=b e^{a T}\int_0^T\mathcal{J}_{\hat{u}}^\top e^{-a t}\,\mathrm{d}t\, \frac{\partial L}{\partial x(T)}\\
&=I\left[D_{\hat{u}}x(T), \mathcal{J}_{\hat{u}}^\top\right]\frac{\partial L}{\partial x(T)}\,.
\end{split}
\label{eq:Lgradient_linear}
\end{equation}
Invoking Eq.~\eqref{eq:induced_descent}, the induced gradient update in $\hat{u}(t;\boldsymbol{\theta}^{(n)})$ is
\begin{equation}
\begin{split}
\hat{u}(t;\boldsymbol{\theta}^{(n+1)})&=\hat{u}(t;\boldsymbol{\theta}^{(n)})-\eta \mathcal{J}_{\hat{u}}b e^{a T}\int_0^T\mathcal{J}_{\hat{u}}^\top e^{-a t}\,\mathrm{d}t\, \frac{\partial L}{\partial x(T)}\\
&=\hat{u}(t;\boldsymbol{\theta}^{(n)})-\eta \mathcal{J}_{\hat{u}}I\left[D_{\hat{u}}x(T), \mathcal{J}_{\hat{u}}^\top\right]\frac{\partial L}{\partial x(T)}\,.
\end{split}
\label{eq:induced_descent_2}
\end{equation}
Based on Eq.~\eqref{eq:induced_descent_2}, we define the weighted total change in $\hat{u}(t;\boldsymbol{\theta}^{(n)})$ for linear dynamics \eqref{eq:dyn_sys_simple_time} as
\begin{align}
\begin{split}
\Delta \hat{U}^{(n)}&\coloneqq \int_0^T\hat{u}(t;\boldsymbol{\theta}^{(n+1)})e^{-a t}\,\mathrm{d}t-\int_0^T\hat{u}(t;\boldsymbol{\theta}^{(n)})e^{-a t}\,\mathrm{d}t\\
&=-\eta^{-1}b^{-1}e^{-aT}\|\Delta {\boldsymbol{\theta}^{(n)}}\|_2^2 \left(\frac{\partial L}{\partial x(T)}\right)^{-1}\,,
\end{split}
\label{eq:delta_U_n}
\end{align}
where $\Delta \boldsymbol{\theta}^{(n)}$ can be directly evaluated with standard backpropagation methods, thus avoiding the evaluation of the Jacobian elements $(\mathcal{J}_{\hat{u}})_i=\partial \hat{u}/\partial \theta_i^{(n)}$ in Eq.~\eqref{eq:induced_descent}.

Figure~\ref{fig:learning_u_GD_vs_Adam} shows a comparison of learned control signals $\hat{u}(t;\boldsymbol{\theta}^{(n)})$ for linear dynamics \eqref{eq:dyn_sys_simple_time} with $x_0=0$ and $T=a=b=x^*=1$. The neural network that we use in this example consists of 2 hidden layers with 6 ELU neurons each. Weights and biases are initialized to values of 0.1. Steepest descent fails to approximate the OC signal [see Fig.~\ref{fig:learning_u_GD_vs_Adam}(a)], whereas Adam is able to approach the optimal control function [see Fig.~\ref{fig:learning_u_GD_vs_Adam}(b)]. We again emphasize that the learning loss is proportional to the squared difference between reached state and target state $(x(T)-x^*)^2$ [see Eq.~\eqref{eq:simple_loss}]. Learning the OC solution is not a direct learning target. Still, in the discussed example, learning OC is possible with the Adam optimizer. As described by Eqs.~\eqref{eq:induced_descent}--\eqref{eq:induced_descent_2}, a gradient descent in $\boldsymbol{\theta}^{(n)}$ may induces a gradient update in $\hat{u}(t;\boldsymbol{\theta}^{(n)})$. Because the Jacobian $\mathcal{J}_{\hat{u}}$ is usually not straightforward to evaluate, we use $\Delta \hat{U}^{(n)}$ to study if the linearization of $\hat{u}(t;\boldsymbol{\theta}^{(n)}+\Delta \boldsymbol{\theta}^{(n)})$ in $\Delta \boldsymbol{\theta}^{(n)}$ is justified. Figure~\ref{fig:learning_u_GD_vs_Adam}(c) shows that $\Delta \hat{U}^{(n)}=\int_0^T\hat{u}(t;\boldsymbol{\theta}^{(n+1)})e^{-a t}\,\mathrm{d}t-\int_0^T\hat{u}(t;\boldsymbol{\theta}^{(n)})e^{-a t}\,\mathrm{d}t$ (red squares) is well-described by $-\eta^{-1}b^{-1}e^{-aT}\|\Delta {\boldsymbol{\theta}^{(n)}}\|_2^2 \left(\frac{\partial L}{\partial x(T)}\right)^{-1}$ (solid red line). The evolution of $\Delta \hat{U}^{(n)}$ under Adam (solid blue line) undergoes oscillations.
\subsection{Control energy regularization}
The induced gradient update in the control input $\hat{u}(t;\boldsymbol{\theta}^{(n)})$ and the evolution of the control energy $E_T[\hat{\mathbf{u}}]$ during training are connected via
\begin{equation}
\begin{split}
E_T[\hat{\mathbf{u}}^{(n+1)}]&=E_T[\hat{\mathbf{u}}^{(n)}]+\int_0^T {\hat{\mathbf{u}}(t;\boldsymbol{\theta}^{(n)})}^\top\mathcal{J}_{\hat{\mathbf{u}}^{(n)}} \,\mathrm{d}t\Delta\boldsymbol{\theta}^{(n)}\\
&+\frac{1}{2}{\Delta \boldsymbol{\theta}^{(n)}}^\top\int_0^T\mathcal{J}_{\hat{\mathbf{u}}^{(n)}}^\top \mathcal{J}_{\hat{\mathbf{u}}^{(n)}}\,\mathrm{d}t \Delta \boldsymbol{\theta}^{(n)}\\
&=E_T[\hat{\mathbf{u}}^{(n)}]-\eta \int_0^T {\hat{\mathbf{u}}(t;\boldsymbol{\theta}^{(n)})}^\top\mathcal{J}_{\hat{\mathbf{u}}^{(n)}} \,\mathrm{d}t\nabla_{\boldsymbol{\theta}^{(n)}}L+\mathcal{O}(\eta^2)\\
&=E_T[\hat{\mathbf{u}}^{(0)}]-\eta \sum_{i=0}^n \int_0^T {\hat{\mathbf{u}}(t;\boldsymbol{\theta}^{(n)})}^\top\mathcal{J}_{\hat{\mathbf{u}}^{(n)}} \,\mathrm{d}t\nabla_{\boldsymbol{\theta}^{(i)}}L+\mathcal{O}(\eta^2)
\,.
\end{split}
\label{eq:energy_evolution}
\end{equation}
Using the identity $\nabla_{\boldsymbol{\theta}^{(n)}} E_T[\hat{\mathbf{u}}^{(n)}]=\int_0^T\mathcal{J}_{\hat{\mathbf{u}}^{(n)}}^\top \hat{\mathbf{u}}(t;\boldsymbol{\theta}^{(n)})\,\mathrm{d}t$, we obtain
\begin{equation}
E_T[\hat{\mathbf{u}}^{(n+1)}]=E_T[\hat{\mathbf{u}}^{(0)}]-\eta \sum_{i=0}^n (\nabla_{\boldsymbol{\theta}^{(i)}} E_T[\hat{\mathbf{u}}^{(i)}])^\top\nabla_{\boldsymbol{\theta}^{(i)}}L+\mathcal{O}(\eta^2)
\,.
\label{eq:energy_evolution_grad}
\end{equation}
\begin{figure}
    \centering
    \includegraphics[width=\textwidth]{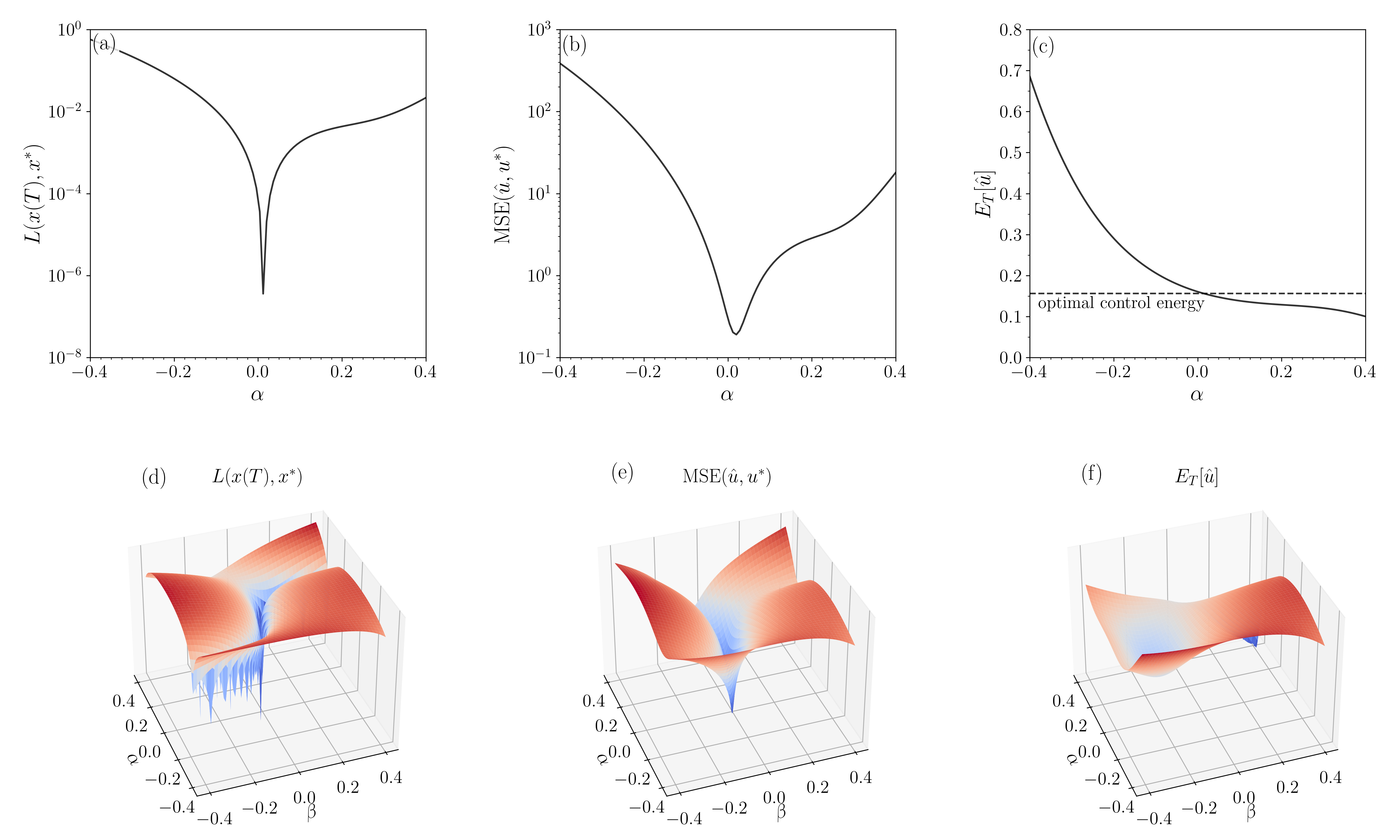}
    \caption{Loss landscapes for training with Adam. Around a local optimum $\boldsymbol{\theta}^*$ obtained after $10^3$ training epochs, we set the neural-network parameters $\boldsymbol{\theta}=\boldsymbol{\theta}^*+\alpha\boldsymbol{\delta}+\beta\boldsymbol{\eta}$. (a,d) The loss $L(x(T),x^*)$ [see Eq.~\eqref{eq:nn_loss}] for $\alpha\in [-0.4,0.4]$, $\beta=0$ (a) and $\alpha,\beta\in[-0.4,0.4]$ (b). (b,e) The mean-squared error associated with the difference between $\hat{u}(t;\boldsymbol{\theta})$ and $u^*$, $\mathrm{MSE}(\hat{u},u^*)$ [see Eq.~\eqref{eq:mse}], as a function of $\alpha\in [-0.4,0.4]$ (b) and $\alpha,\beta\in[-0.4,0.4]$ (e). (c,f) The control energy $E_T[\hat{u}]$ [see Eq.~\eqref{eq:control_energy}] as a function of $\alpha\in [-0.4,0.4]$ (c) and $\alpha,\beta\in[-0.4,0.4]$ (f). The optimal control energy, $1/(e^2-1)$, is indicated by a dashed black line in panel (c). The neural network that we use in this example consists of 2 hidden layers with 6 ELU neurons each. Parameters are set to $x_0=0$ and $T=a=b=x^*=1$. We used Adam to train the neural network and set the learning rate to $\eta=0.15$. Initial weights and biases were set to $0.1$.}
    \label{fig:loss_landscape_6_elus_adam}
\end{figure}
\begin{figure}
    \centering
    \includegraphics[width=\textwidth]{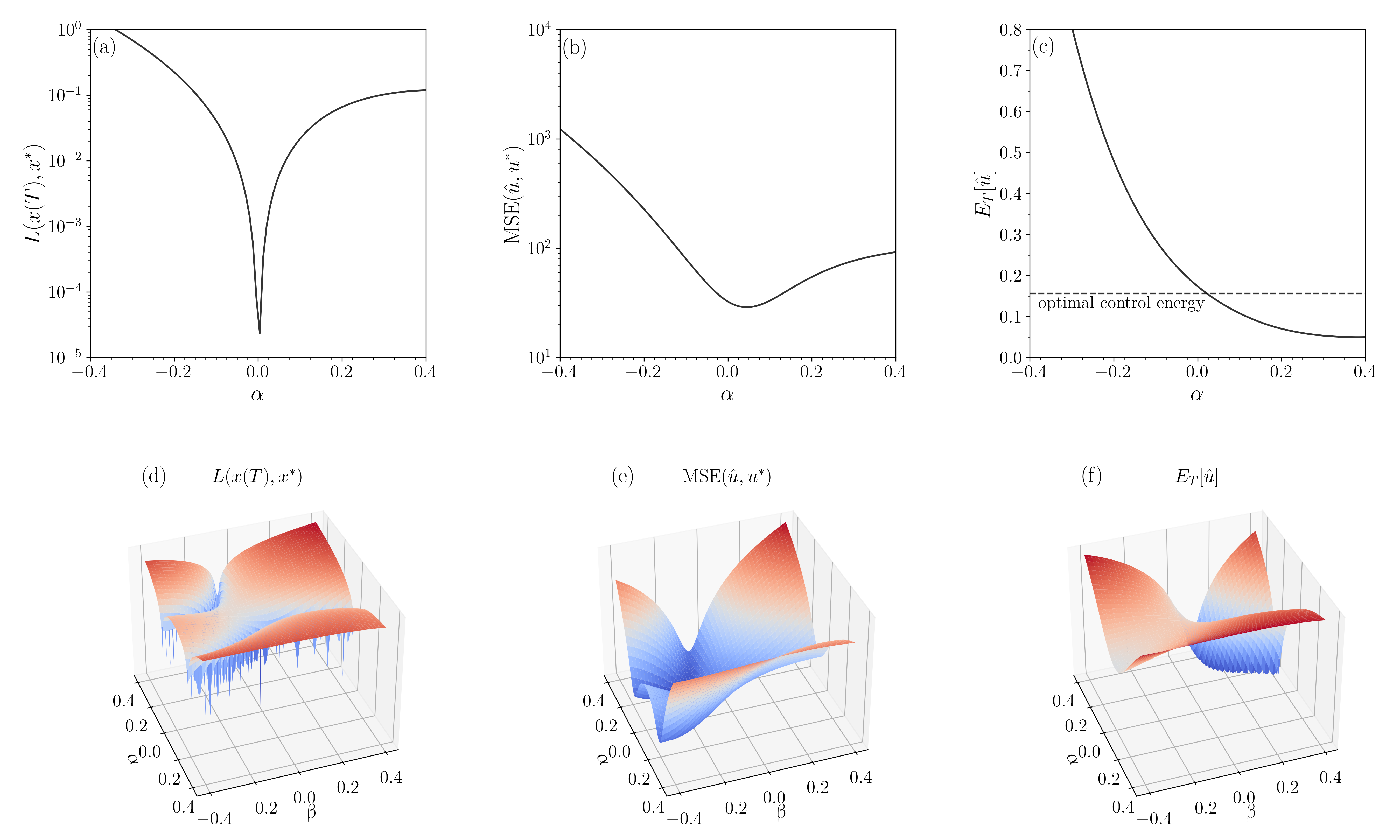}
    \caption{Loss landscapes for training with steepest descent. Around a local optimum $\boldsymbol{\theta}^*$ obtained after $10^3$ training epochs, we set the neural-network parameters $\boldsymbol{\theta}=\boldsymbol{\theta}^*+\alpha\boldsymbol{\delta}+\beta\boldsymbol{\eta}$. (a,d) The loss $L(x(T),x^*)$ [see Eq.~\eqref{eq:nn_loss}] as a function of $\alpha\in [-0.4,0.4]$, $\beta=0$ (a) and $\alpha,\beta\in[-0.4,0.4]$ (b). (b,e) The mean-squared error associated with the difference between $\hat{u}(t;\boldsymbol{\theta})$ and $u^*$, $\mathrm{MSE}(\hat{u},u^*)$ [see Eq.~\eqref{eq:mse}], as a function of $\alpha\in [-0.4,0.4]$ (b) and $\alpha,\beta\in[-0.4,0.4]$ (e). (c,f) The control energy $E_T[\hat{u}]$ [see Eq.~\eqref{eq:control_energy}] as a function of $\alpha\in [-0.4,0.4]$ (c) and $\alpha,\beta\in[-0.4,0.4]$ (f). The optimal control energy, $1/(e^2-1)$, is indicated by a dashed black line in panel (c). The neural network that we use in this example consists of 2 hidden layers with 6 ELU neurons each. Parameters are set to $x_0=0$ and $T=a=b=x^*=1$. We used steepest descent to train the neural network and set the learning rate to $\eta=0.15$. Initial weights and biases were set to $0.1$.}
    \label{fig:loss_landscape_6_elus_sgd}
\end{figure}
Equation~\eqref{eq:energy_evolution} shows that, up to terms of higher order, a gradient-descent update in $\boldsymbol{\theta}$ is associated with a control energy update that is equal to $(\nabla_{\boldsymbol{\theta}^{(n)}} E_T[\hat{\mathbf{u}}^{(n)}])^\top$ multiplied by $\Delta \boldsymbol{\theta}^{(n)}=-\eta \nabla_{\boldsymbol{\theta}^{(n)}}L$. Let $\alpha$ denote the angle between $\nabla_{\boldsymbol{\theta}^{(n)}} E_T[\hat{\mathbf{u}}^{(n)}]$ and $\nabla_{\boldsymbol{\theta}^{(n)}}L$. If $\cos(\omega)>0$ (\ie, if $-\pi/2<\omega<\pi/2$), a gradient update in $\boldsymbol{\theta}^{(n)}$ is associated with a decrease in control energy. 

Table~\ref{tab:gradient_descents} provides an overview of the gradient updates associated with the neural-network parameters $\boldsymbol{\theta}^{(n)}$, control function $\mathbf{u}(t;\boldsymbol{\theta}^{(n)})$, and control energy $E_T[\hat{\mathbf{u}}^{(n+1)}]$.
\begin{table}[htb]
{\centering
\renewcommand*{\arraystretch}{1.2}
\small
\begin{tabular}{
>{\raggedright\arraybackslash} m{15em}
>{\raggedright\arraybackslash} m{22em}
}\toprule
Neural-network parameters & $\boldsymbol{\theta}^{(n+1)}=\boldsymbol{\theta}^{(n)}-\eta \nabla_{\boldsymbol{\theta}^{(n)}}L$ \\[1pt]
Control function & $\hat{\mathbf{u}}(t;\boldsymbol{\theta}^{(n+1)})=\hat{\mathbf{u}}(t;\boldsymbol{\theta}^{(n)})-\eta \mathcal{J}_{\hat{\mathbf{u}}} \nabla_{\boldsymbol{\theta}^{(n)}}L$ \\[1pt]
Control energy & $E_T[\hat{\mathbf{u}}^{(n+1)}]=E_T[\hat{\mathbf{u}}^{(n)}]-\eta(\nabla_{\boldsymbol{\theta}^{(n)}} E_T[\hat{\mathbf{u}}^{(n)}])^\top\nabla_{\boldsymbol{\theta}^{(n)}}L$ \\[1pt] \bottomrule
\end{tabular}
\vspace{1mm}}
\caption{Overview of gradient updates of neural-network parameters, control function, and control energy.}
\label{tab:gradient_descents}
\end{table}

Using random projections of the loss function $L$~\cite{DBLP:conf/nips/Li0TSG18} can help provide geometric intuition for the interplay between explicit minimization of $L$ and implicit regularization of the control energy $E_T[\hat{u}]$. To obtain two and three-dimensional projections of the 51-dimensional parameter space of the employed neural networks, we set the neural-network parameters $\boldsymbol{\theta}=\boldsymbol{\theta}^*+\alpha\boldsymbol{\delta}+\beta\boldsymbol{\eta}$ around a local optimum $\boldsymbol{\theta}^*$. Here, $\alpha,\beta\subseteq\mathbb{R}$ denote scaling parameters and $\boldsymbol{\delta},\boldsymbol{\eta}\sim\mathcal{N}(0,\mathds{1}_{N\times N})$ are random Gaussian vectors.

For an Adam-based optimization, Fig.~\ref{fig:loss_landscape_6_elus_adam} shows two and three-dimensional random projections of $L(x(T),x^*)$, $E_T[\hat{u}]$, and the mean-squared error
\begin{equation}
\mathrm{MSE}(u^*,\hat{u})=\frac{1}{M}\sum_{i=1}^M [u^*(t_i)-\hat{u}(t_i)]^2\approx \int_0^T [u^*(t)-\hat{u}(t)]^2\,\mathrm{d}t\,,
\label{eq:mse}
\end{equation}
quantifying the deviation of $\hat{u}$ from the optimum $u^*$. Here, $M$ is the number of timesteps and $t_i = i\,T/M$. In the random projection shown in Fig.~\ref{fig:loss_landscape_6_elus_adam}, we observe that Adam produces a local optimum that is associated with a small loss, $\mathrm{MSE}(u^*,\hat{u})$, and control energy. In comparison with the results obtained with steepest descent (see Fig.~\ref{fig:loss_landscape_6_elus_sgd}), we observe that Adam produces sharper local optima. Figure~\ref{fig:loss_landscape_6_elus_sgd} also shows that steepest descent produces two ``valleys'' of local optima separated by a plateau. Escaping from a local optimum located in one valley to another optimum in the second valley may be challenging as even larger learning rates may not be able to overcome the plateau. Additional loss projections for neural networks with fewer parameters are provided in Appendix~\ref{app:loss_surfaces}.
\subsection{Constant optimal control approximation}
\label{sec:const_contr_approx}
\begin{figure}
    \centering
    \includegraphics[width=\textwidth]{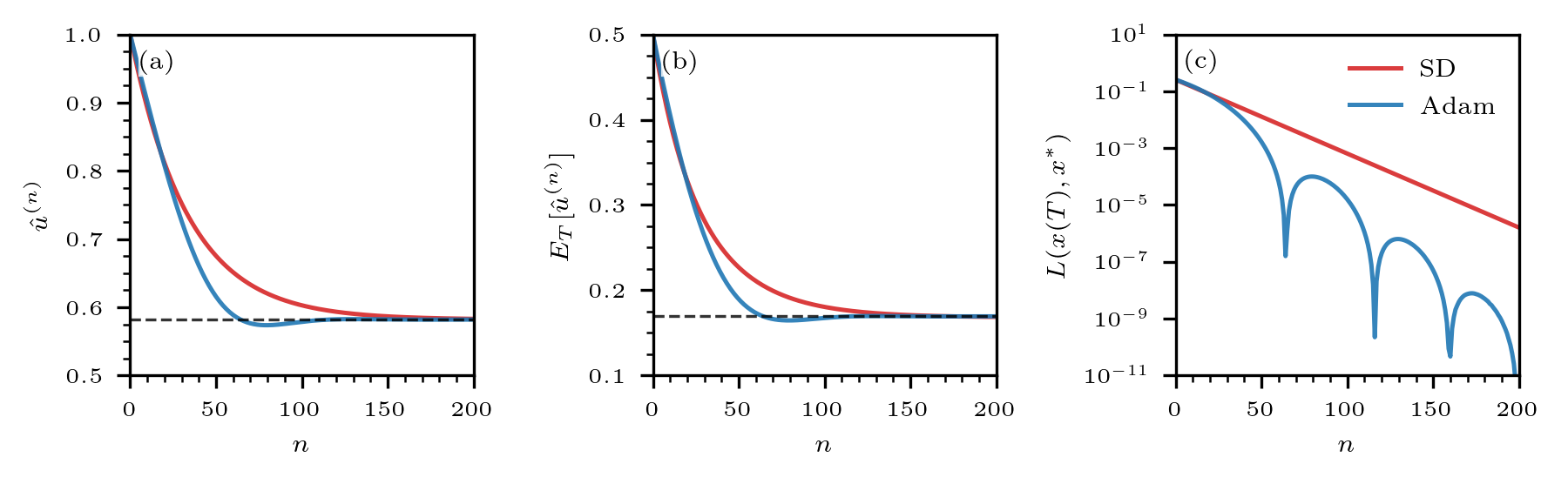}
    \caption{Approximating optimal control by a constant. (a) The constant control input $\hat{u}(t;\boldsymbol{\theta}^{n})\equiv c^{(n)}$ as a function of the number of training epochs $n$. The dashed black line indicates the optimum $c^{(\infty)}=1/(e-1)\approx 0.58$. (b) The control energy $E_T[\hat{u}^{(n)}]$ as a function of $n$. In the limit $n\rightarrow\infty$, the control energy approaches $1/2 (e-1)^{-2}\approx 0.17$. (c) The MSE loss $L(x(T),x^*)$ as a function of $n$. Solid red and blue lines represent steepest descent and Adam solutions, respectively. We set the learning rate $\eta=10^{-2}$.}
    \label{fig:constant_approx}
\end{figure}
In situations where the OC function is expected to only vary modestly in time, or when no good guess of the OC solution is available, one may use a constant control approximation as a baseline. For such a baseline, the underlying neural network consists of a bias term (\ie, $\hat{u}(t;\boldsymbol{\theta}^{(n)})\equiv c^{(n)}$) and Eq.~\eqref{eq:induced_descent} becomes
\begin{equation}
c^{(n+1)}=c^{(n)}+\Delta c^{(n)}\,,
\label{eq:induced_descent_constant}
\end{equation}
where $\Delta c^{(n)}=-\eta \frac{\mathrm{d} L}{\mathrm{d}c^{(n)}}$ and
\begin{equation}
L=\frac{1}{2}\left[\int_0^1 e^{-t+1} c^{(n)}\,\mathrm{d}t-1\right]^2=\frac{1}{2}\left[c^{(n)}(e-1)-1\right]^2\,.
\end{equation}
In the limit $n\rightarrow\infty$, the control function approaches $c^*=1/(e-1)$.

In each gradient-descent step, the control energy changes according to
\begin{equation}
E_T[\hat{u}^{(n+1)}]=E_T[\hat{u}^{(n)}]-\eta c^{(n)} T\frac{\mathrm{d} L}{\mathrm{d} b^{(n)}}\,.
\end{equation}

For linear dynamics \eqref{eq:dyn_sys_simple_time} with $x_0=0$ and $T=a=b=x^*=1$, we have 
\begin{equation}
\begin{split}
E_T[\hat{u}^{(n+1)}]&=E_T[\hat{u}^{(n)}]-\eta c^{(n)} \int_0^1 e^{-t+1}\,\mathrm{d}t\, \left[\int_0^1 e^{-t+1} c^{(n)}\,\mathrm{d}t+1\right]\\
&=E_T[\hat{u}^{(n)}]-\eta c^{(n)} (e-1) \left[{c^{(n)}} (e-1)-1\right]\,.
\end{split}
\end{equation}

Figure~\ref{fig:constant_approx} shows the evolution of $\hat{u}(t;\boldsymbol{\theta}^{(n)})\equiv c^{(n)}$, $E_T[\hat{u}^{(n)}]$, and $L(x(T),x^*)$. Solid red and blue lines represent solutions that were obtained with steepest descent and Adam, respectively. Dashed black lines in Fig.~\ref{fig:constant_approx}(a,b) indicate the OC solutions. The convergence behavior of Adam is non-monotonic. Between 60 and 70 training epochs, Adam achieves control energy values smaller than $E_T[\hat{u}^{(\infty)}]=1/2(e-1)^{-2}$, owing to an increase in the loss $L(x(T),x^*)$. The ratio between $E_T[\hat{u}^{(\infty)}]=1/2(e-1)^{-2}$ and the OC energy is 1.08, indicating that a constant control approximation achieves a control energy that is close to that of the corresponding OC solution.

The constant control baseline that we derived in this section also appears to be close to possible local optima that are learned by NODEC. For example, for certain learning rates and initial values of neural-network parameters, Adam approaches solutions that are similar to the constant baseline [see Fig.~\ref{fig:time_dependent_control}(b,e)]. We also observe a similar behavior for steepest descent learning [see Fig.~\ref{fig:learning_u_GD_vs_Adam}(a)]. Also one of the loss projections in Appendix~\ref{app:loss_surfaces} shows that steepest descent approaches a local optimum that corresponds to a constant control approximation. A solution that is closer to OC and which was not found by steepest descent is also visible in the shown projections.
\section{Discussion}
\label{sec:discussion}
Optimal control problems arise in various applications and can be solved analytically often only under special assumptions (\eg, for linear systems~\cite{yan2012controlling}). For non-linear control problems, different direct and indirect numerical methods have been developed to approximate OC solutions. The loss function underlying OC problems consists of a final cost (\eg, difference between reached and target state) and an integrated cost (\eg, control energy). To achieve a certain tradeoff between these two cost contributions, one has to find suitable Lagrange multipliers [see Eq.~\eqref{eq:oc_loss}], which might prove challenging in practice. Neural ODE controllers provide an alternative to standard direct and indirect OC solvers in that they are able to implicitly learn near-optimal control functions~\cite{asikis2022neural,bottcher2022ai} if the underlying hyperparameters are chosen appropriately. The problem of optimizing Lagrange multipliers in OC loss functionals can be recast into a hyperparameter optimization problem, with the goal being that the basin of attraction of the desired OC solution in the underlying loss landscape is sufficiently large.

To study the ability of neural ODE controllers to perform implicit control energy regularization, we have analyzed the influence of backpropagation protocols, parameter initialization, optimizers, and activation functions on the learned control solutions. Using analytical and numerical arguments, we were able to identify different features that can help neural ODE control designers to efficiently perform hyperparameter optimization and learn near-optimal control solutions.

First, truncated backpropagation through time (TBPTT) has been shown to be computationally more efficient in terms of iterations per unit time than its untruncated counterpart. Although TBPTT protocols only provide an approximation of the actual gradient update, our results suggest that they can achieve loss values and control solutions that are similar to those obtained with untruncated gradient updates. Future work may evaluate the computational advantages of TBPTT over BPTT in larger problems.

Second, a useful starting point for certain control problems may be a constant controller that can be represented by a bias term. Constant optimal control approximations appeared as local optima in deeper architectures that we used in our numerical experiments. For high-dimensional control tasks studied in earlier work~\cite{bottcher2022ai}, NODEC also learned a constant optimal control approximation that provided better performance than solutions found by an adjoint-gradient method (\ie, an indirect OC solver). Increasing the width and depth of the underlying ANN can substantially improve the performance of NODEC in implicitly learning near-optimal control solutions as shown in Sec.~\ref{sec:nn_depth_width}. Between four to six hidden fully-connected layers and about five to ten neurons per hidden layer have shown good performance in different control tasks. In accordance with earlier numerical work~\cite{bottcher2022ai}, our results also highlight the importance of appropriate initialization protocols. Large weight and bias values may often be associated with loss values that are far away from the basin of attraction of a near-optimal control solution. In this context it may be worthwhile to study combinations of neural-network parameter initialization protocols and momentum schedulers~\cite{sutskever2013importance}.

Third, adapative gradient methods like Adam have been shown to be able to learn near-optimal control functions in problems where steepest descent fails to do so. These results therefore contribute to the discussion of the advantages of adaptive optimizers over standard gradient descent techniques~\cite{hardt2016train,wilson2017marginal,choi2020on}. Two and three dimensional loss projections provide insights into hyperparameter-dependent geometric features of the loss and control energy landscapes and may reveal the existence of better nearby local optima.

There are different possibilities for future work. Studying the ability of neural ODE controllers to implicitly learn control functions for loss functionals different from Eq.~\eqref{eq:oc_loss} (``generalized implicit regularization''). It would also be interesting to use dynamical systems approaches to study the behavior of adaptive optimizers akin to our steepest-descent analysis in Sec.~\ref{sec:constant_control}. Instead of using ANNs to approximate control functions, control problems can function as a tool to better understand optimization dynamics within deep ANNs that are otherwise mainly studied in terms of image classification.
\bibliographystyle{siamplain}
\bibliography{refs.bib}

\newpage
\appendix 
\section{Moving particle subject to friction}
\label{app:moving_particle}
\begin{figure}
    \centering
    \includegraphics[width=\textwidth]{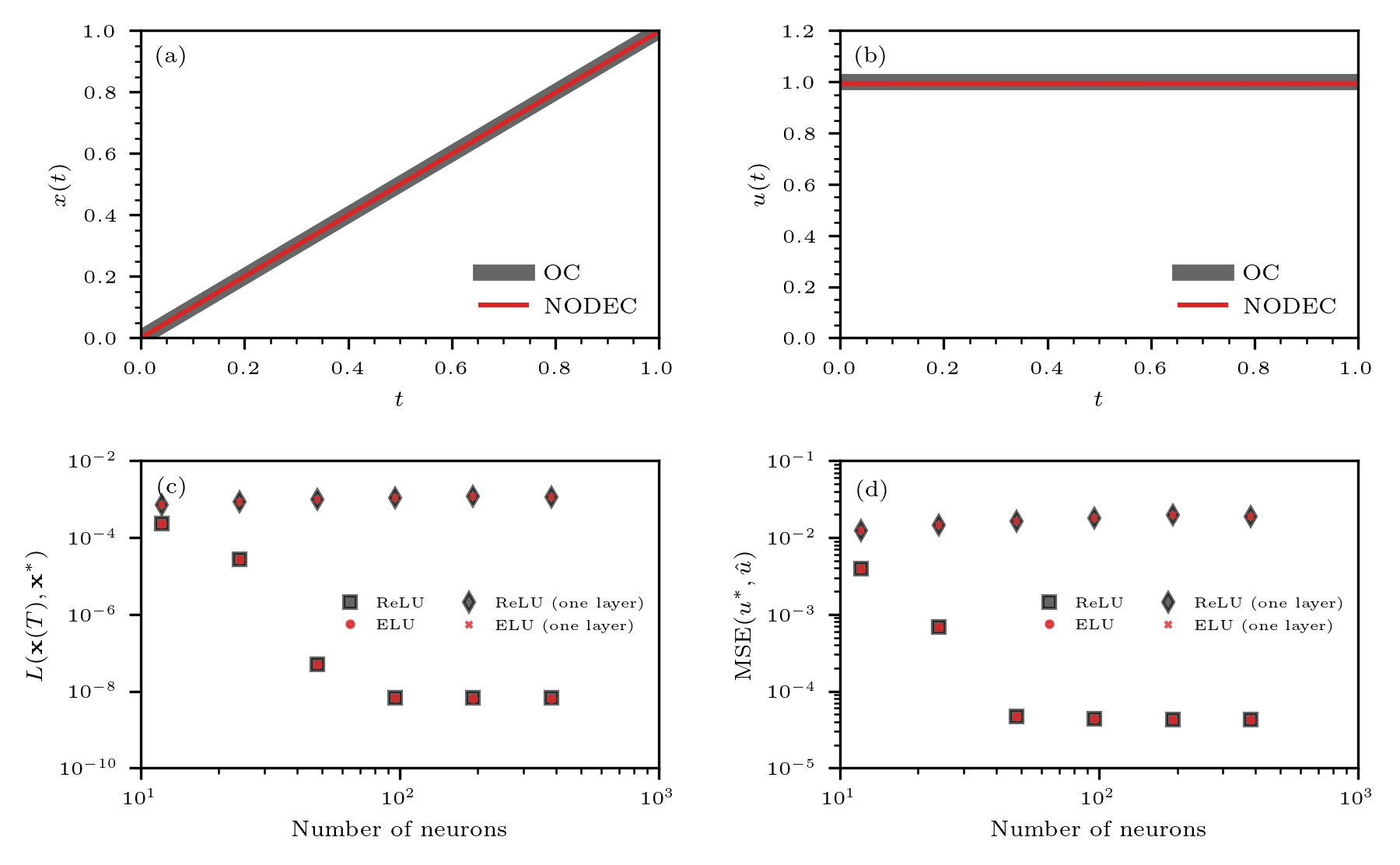}
    \caption{Controlling a moving particle that is subject to friction. (a,b) Evolution of the state $x(t)$ and control function $u(t)$. Optimal control and NODEC-based solutions are shown in grey and red, respectively. (c,d) Loss $L(\mathbf{x}(T),\mathbf{x}^*)=\|\mathbf{x}(t)-\mathbf{x}^*\|_2^2$ and $\mathrm{MSE}(\mathbf{u}^*,\hat{\mathbf{u}})$ for different activations and architectures as a function of the number of neurons.}
    \label{fig:moving_particle}
\end{figure}
The goal is to minimize
\begin{equation}
W[\mathbf{x},u]=\int_0^1 v(t)u(t) \,\mathrm{d}t
\label{eq:work}
\end{equation}
subject to
\begin{align}
\begin{split}
\dot{x}(t)&=v(t)\\
\dot{v}(t)&=-v(t)+u(t)\\
v(t)&\geq 0\\
0\leq &u(t)\leq 2\\
(x(0),v(0))&=(0,1)\\
(x(1),v(1))&=(1,1)\,,
\end{split}
\end{align}
where $\mathbf{x}(t)=(x(t),v(t))^\top$~\cite{gong2006pseudospectral}. This control problem describes the movement of a particle under friction from $x(0)=0$ to $x(1)=1$ minimizing the work done. The control input $u(t)$ corresponds to a force that acts on the particle and the optimal control in this example is $u^*(t)=1$.

To solve this control problem with NODEC, we represent $u(t)$ by a neural network $\hat{u}(t;\boldsymbol{\theta})$ with $H$ layers and $N_H$ neurons per layer. We train NODEC using Adam and the MSE loss and set $L(\mathbf{x}(T),\mathbf{x}^*)=\|\mathbf{x}(t)-\mathbf{x}^*\|_2^2$. That is, we not explicitly account for control bounds and $W[\mathbf{x},u]$).
\begin{figure}
    \centering
    \includegraphics[width=\textwidth]{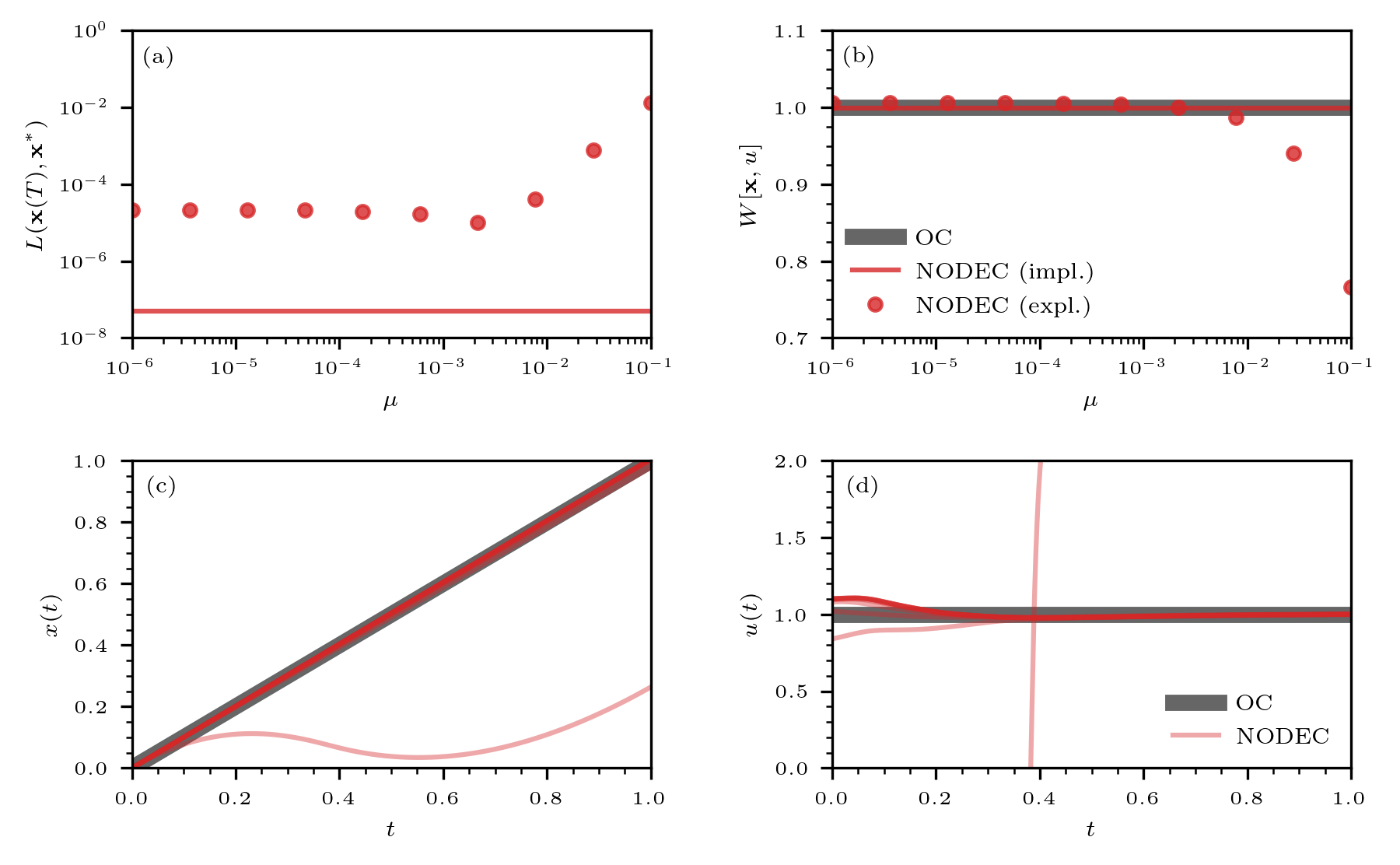}
    \caption{Lagrange multiplier optimization. (a,b) The loss $L(\mathbf{x}(T),\mathbf{x}^*)=\|\mathbf{x}(T)-\mathbf{x}^*\|_2^2$ and work $W[\mathbf{x},u]$ [see Eq.~\eqref{eq:work}] for different values of $\mu$ [see Eq.~\eqref{eq:loss_2}]. Red disks are associated with a direct optimization of both $L(\mathbf{x}(T),\mathbf{x}^*)$ and $W[\mathbf{x},u]$ while solid red lines indicate implicit regularization solutions. (c,d) Different values of $\mu$ are associated with different NODEC-based solutions of $x(t)$ and $\hat{u}(t;\boldsymbol{\theta})$ (solid red lines). As a reference, we also show the corresponding OC solutions (solid grey line).}
    \label{fig:moving_particle_2}
\end{figure}
Figure~\ref{fig:moving_particle}(a,b) shows the evolution of $x(t)$ and $u(t)$. Optimal control and NODEC solutions are represented by solid grey and red lines, respectively. The neural network that we use in this example has $H=64$ layers with $N_H=6$ ELU neurons per layer. Bias terms are included in every layer.

To study the effect of different neural network structures and activations on the learning performance, we first set the number of neurons per layer $N_H=6$ and vary the number of layers $H$ from 2 to 64. All weights and biases are initially set to a value of $10^{-2}$. We perform numerical experiments for both ELU and ReLU activations. We train NODEC for 100 epochs using Adam with a learning rate $\eta=0.5\times 10^{-2}$, and we evaluate the best model. We show the corresponding loss and MSE values after training in Fig.~\ref{fig:moving_particle}(c,d). For $H\geq 8$, the loss approaches values smaller than $10^{-7}$ and the MSE drops to values below $10^{-4}$. For one layer and varying $N_H$, the neural network is not able approximate the OC solution. Our results are almost identical for both ReLU and ELU activations.

In addition to the prior numerical experiment, we now also directly account for the additional loss term $W[\mathbf{x},u]$ and optimize
\begin{equation}
L(\mathbf{x}(T),\mathbf{x}^*)+\mu W[\mathbf{x},u]\,,
\label{eq:loss_2}
\end{equation}
where $\mu$ is a Lagrange multiplier that models the influence of $W[\cdot]$ on the total loss. To initialize weights without additional parameter optimization, we use the Kaiming uniform initializer \cite{he2015delving}, and we initially set all biases to a value of $10^{-2}$. The neural network that we use to optimize Eq.~\eqref{eq:loss_2} consists of $H=8$ fully connected linear layers with $N_H=6$ ELU neurons each. This way of setting up the network architecture ensures that NODEC is able to represent the constant control function. We use Adam with a learning rate of $10^{-1}$ to train the neural network for 100 epochs, and we evaluate the best model. In addition to the uniform weight initialization, the learning rate of $\eta=10^{-1}$ was chosen to model a not fully optimized hyperparameter set. Otherwise, near-optimal solutions would be learned automatically as in the previous example without optimizing the multiplier $\mu$. As shown in Fig.~\ref{fig:moving_particle_2}(a,b), the loss component $L(\mathbf{x}(T),\mathbf{x}^*)$ approaches a minimum for $\mu \approx 2\times 10^{-3}$. The corresponding value of $W[\mathbf{x},u]$ is reasonably close to the optimal solution. However, Fig.~\ref{fig:moving_particle_2}(a) also shows that the implicit regularization of $W[\mathbf{x},u]$ can achieve smaller loss values and similar values of $W[\mathbf{x},u]$. Figure~\ref{fig:moving_particle_2}(c,d) shows different NODEC solutions (solid red lines) that are associated with different multiplier values $\mu$. Although almost all solutions are aligned with the OC solution in terms of the evolution of $x(t)$, variations in $\mu$ produce visible deviations of $\hat{u}(t;\boldsymbol{\theta})$ from $u^*(t)$. 
\section{Architectures for time-dependent control in two dimensions}
\label{app:time_dependent}
\begin{figure}
    \centering
    \includegraphics[width=\textwidth]{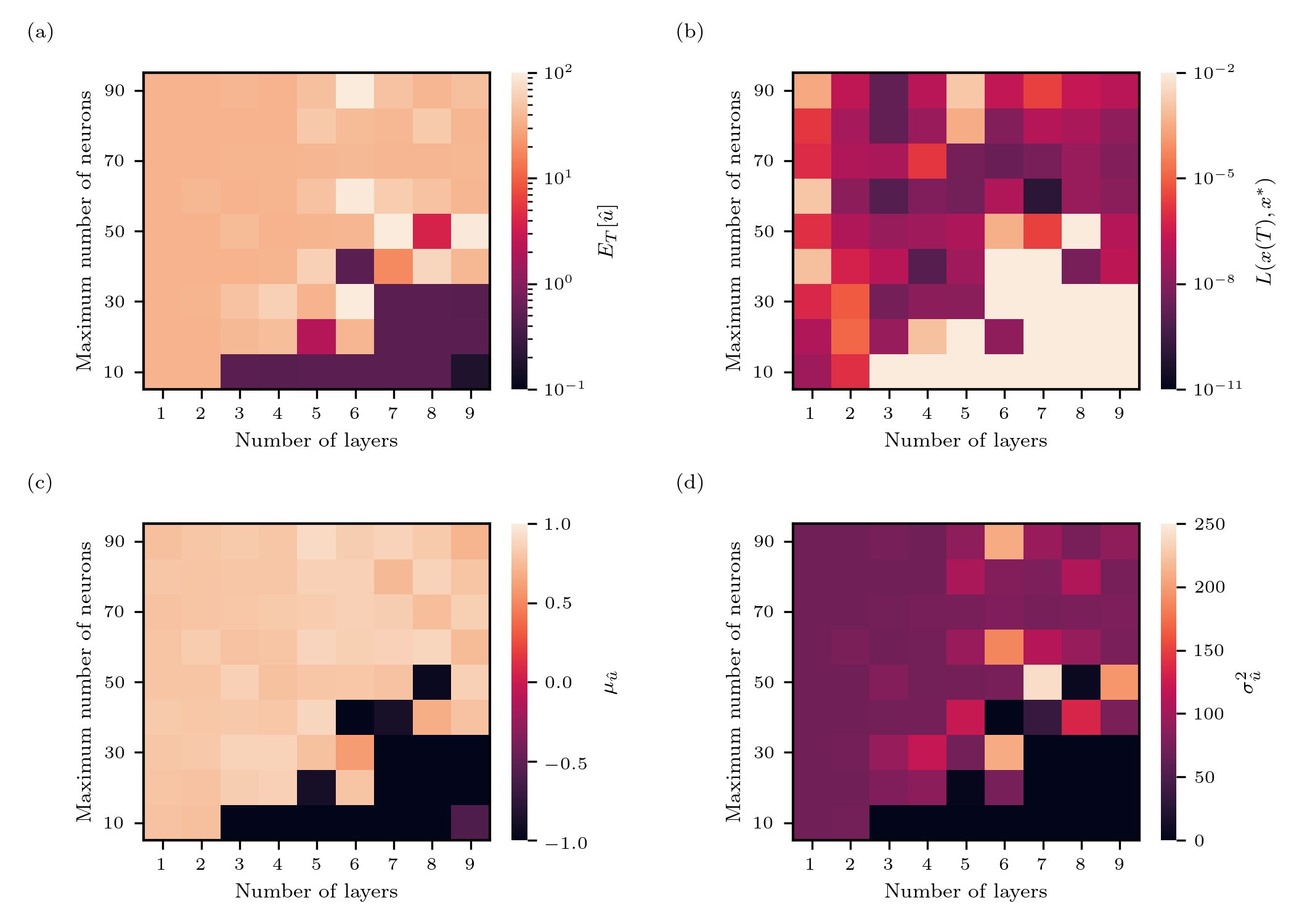}
    \caption{Effect of variations in number of layers and neurons per layer on the ability of NODEC to learn a time-dependent control signal for the two-dimensional flow \eqref{eq:2d_flow}. The underlying architectures use bias terms and leaky ReLU activations in all hidden layers. Each ANN is trained for 500 epochs using Adam and learning rate $\eta=3 \times 10^{-3}$. We set the number of neurons per layer to be the greatest integer less than or equal to the maximum number of neurons (vertical axes) divided by the number of layers (horizontal axes). The optimal control energy is approximately $34$. Heatmaps show the (a) control energy $E_T[\hat{u}]$, (b) loss $L(x(T),x^*)$, (c) mean control signal over time $\mu_{\hat{u}}$, and (d) control signal variance over time, $\sigma^2_{\hat{u}}$.
    }
    \label{fig:depth_vs_width_linear2d}
\end{figure}
For learning a time-dependent control, we use an ANN with leaky ReLU activations, between one to nine layers, and up to 90 neurons. Weights and biases are initially distributed according to $\mathcal{U}(-\sqrt{k},\sqrt{k})$, where $k$ denotes the number of input features of a certain layer. Figure~\ref{fig:depth_vs_width_linear2d}(a) shows that smaller control energies can be achieved as the number of layers increases. For six layers and 20 neurons in total, we observe that the ANN controller is able to implicitly learn a solution with a small loss and a control energy that is close to that of the optimal solution. However, in general, the loss may increase with the number of layers [see Fig.~\ref{fig:depth_vs_width_linear2d}(b)].
Increasing the layers while keeping a low number of neurons seems to affect convergence as most ANNs seem to converge to a non-optimal constant control\textemdash the variance in the bottom right corner of Fig.~\ref{fig:depth_vs_width_linear2d}(d) is close to 0 (darker color).
To summarize, implicit regularization of control energy can be achieved by selecting appropriate hyperparameters (\eg, six layers and 20 neurons in total).
\section{Loss projections}
\label{app:loss_surfaces}

\begin{figure}
    \centering
    \includegraphics[width=\textwidth]{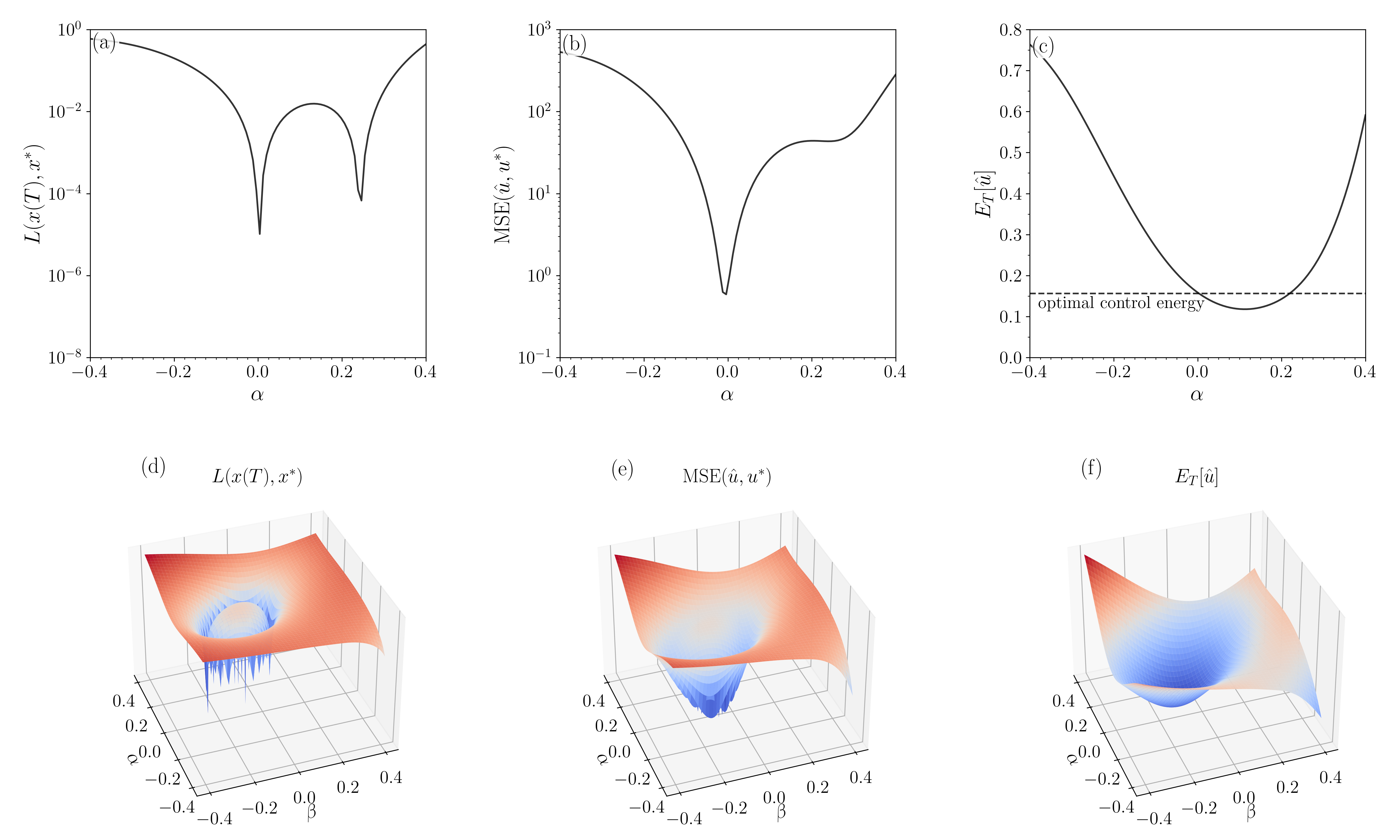}
    \caption{Loss landscapes for training with Adam. Loss landscapes for training with Adam. Around a local optimum $\boldsymbol{\theta}^*$ obtained after $10^3$ training epochs, we set the neural-network parameters $\boldsymbol{\theta}=\boldsymbol{\theta}^*+\alpha\boldsymbol{\delta}+\beta\boldsymbol{\eta}$. (a,d) The loss $L(x(T),x^*)$ [see Eq.~\eqref{eq:nn_loss}] as a function of $\alpha\in [-0.4,0.4]$, $\beta=0$ (a) and $\alpha,\beta\in[-0.4,0.4]$ (b). (b,e) The mean-squared error associated with the difference between $\hat{u}(t;\boldsymbol{\theta})$ and $u^*$, $\mathrm{MSE}(\hat{u},u^*)$ [see Eq.~\eqref{eq:mse}], as a function of $\alpha\in [-0.4,0.4]$ (b) and $\alpha,\beta\in[-0.4,0.4]$ (e). (c,f) The control energy $E_T[\hat{u}]$ [see Eq.~\eqref{eq:control_energy}] as a function of $\alpha\in [-0.4,0.4]$ (c) and $\alpha,\beta\in[-0.4,0.4]$ (f). The optimal control energy, $1/(e^2-1)$, is indicated by a dashed black line in panel (c). The neural network that we use in this example consists of 2 hidden layers with 5 ELU neurons each. Parameters are set to $x_0=0$ and $T=a=b=x^*=1$. We used Adam to train NODEC and we set the learning rate to $\eta=0.1733$. Initial weights and biases are set to $0.1$.}
    \label{fig:loss_landscape_5_elus_adam}
\end{figure}
\begin{figure}
    \centering
    \includegraphics[width=\textwidth]{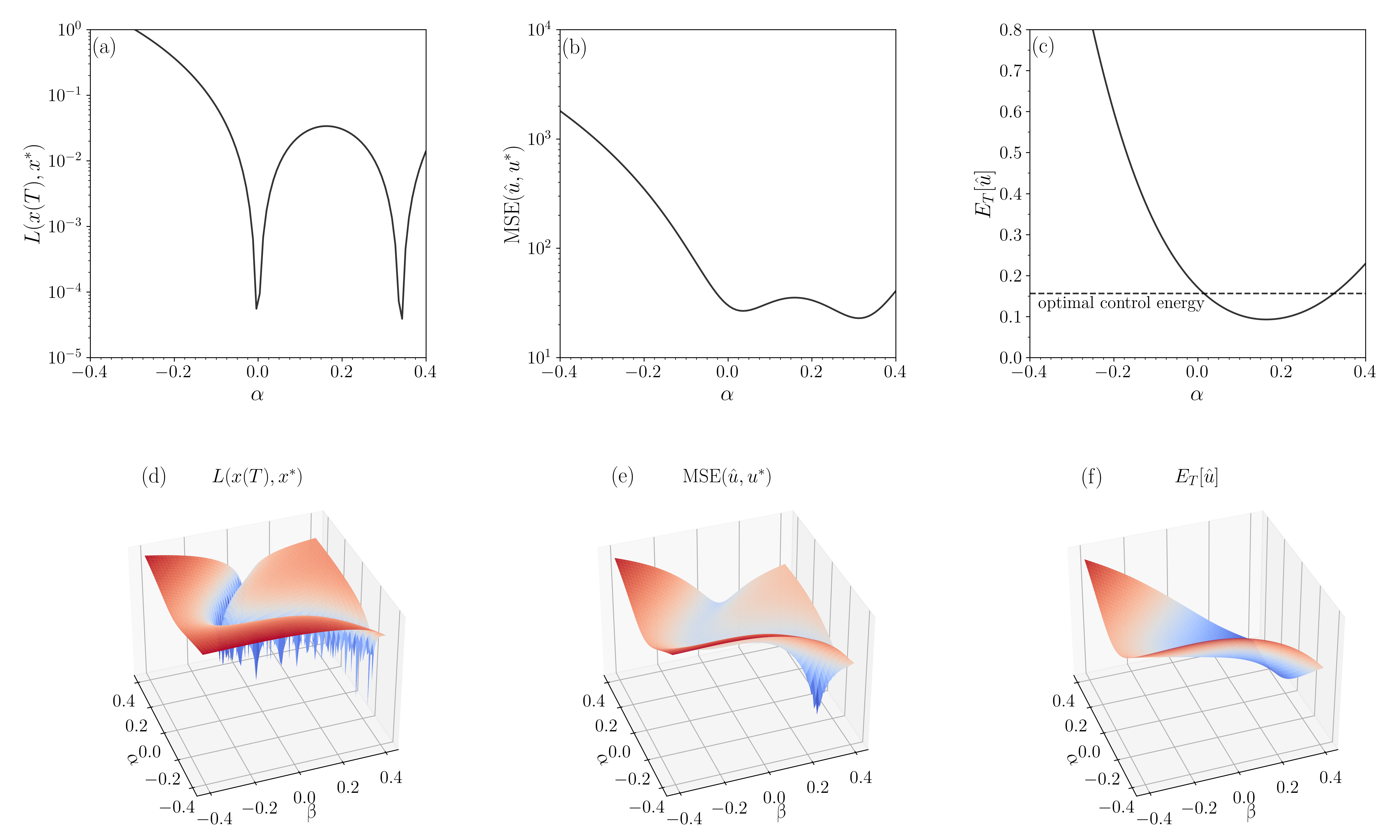}
    \caption{Loss landscapes for training with steepest descent. Loss landscapes for training with Adam. Around a local optimum $\boldsymbol{\theta}^*$ obtained after $10^3$ training epochs, we set the neural-network parameters $\boldsymbol{\theta}=\boldsymbol{\theta}^*+\alpha\boldsymbol{\delta}+\beta\boldsymbol{\eta}$. (a,d) The loss $L(x(T),x^*)$ [see Eq.~\eqref{eq:nn_loss}] as a function of $\alpha\in [-0.4,0.4]$, $\beta=0$ (a) and $\alpha,\beta\in[-0.4,0.4]$ (b). (b,e) The mean-squared error associated with the difference between $\hat{u}(t;\boldsymbol{\theta})$ and $u^*$, $\mathrm{MSE}(\hat{u},u^*)$ [see Eq.~\eqref{eq:mse}], as a function of $\alpha\in [-0.4,0.4]$ (b) and $\alpha,\beta\in[-0.4,0.4]$ (e). (c,f) The control energy $E_T[\hat{u}]$ [see Eq.~\eqref{eq:control_energy}] as a function of $\alpha\in [-0.4,0.4]$ (c) and $\alpha,\beta\in[-0.4,0.4]$ (f). The optimal control energy, $1/(e^2-1)$, is indicated by a dashed black line in panel (c). The neural network that we use in this example consists of 2 hidden layers with 5 ELU neurons each. Parameters are set to $x_0=0$ and $T=a=b=x^*=1$. We used steepest descent to train NODEC and we set the learning rate to $\eta=0.1733$. Initial weights and biases are set to $0.1$.}
    \label{fig:loss_landscape_5_elus_sgd}
\end{figure}

To complement our geometric analysis of implicit energy regularization (see Sec.~\ref{sec:impl_regularization}), we provide additional loss, MSE, and control energy visualizations in Figs.~\ref{fig:loss_landscape_5_elus_adam} and \ref{fig:loss_landscape_5_elus_sgd}. The control task is the same as in Sec.~\ref{sec:impl_regularization}. We aim at controlling the one-dimensional flow \eqref{eq:dyn_sys_simple_time} with $x_0=0$ and $T=a=b=x^*=1$. The neural network that we use here consists of 2 hidden layers with 5 ELU neurons each. Weights and biases are initialized to values of 0.1. The total number of neural-network parameters is 46 instead of 51 as in Sec.~\ref{sec:impl_regularization}.

Figure~\ref{fig:loss_landscape_5_elus_adam} shows two and three-dimensional loss projections around a local optimum found by Adam. The local optimum is associated with a small loss, MSE, and control energy. Other surrounding local optima are not closer to OC solution.

For the same neural network and initial conditions, we observe in Fig.~\ref{fig:loss_landscape_5_elus_sgd} that steepest descent gets stuck in a local optimum that is associated with a larger MSE than the optimum found by Adam. Another optimum that is visible in the projections in Fig.~\ref{fig:loss_landscape_5_elus_sgd} has a significantly smaller MSE and control energy than the one found by steepest descent. However, in the shown example, steepest descent converges towards a constant control approximation (see Sec.~\ref{sec:const_contr_approx}) and not towards the local optimum that is closer to the OC solution.

\end{document}